\documentclass{article}

\usepackage{microtype}
\usepackage{graphicx}
\usepackage{subfigure}
\usepackage{booktabs} 
\usepackage{amsmath}
\usepackage{amsfonts}
\usepackage{paralist}
\usepackage[table,xcdraw]{xcolor}
\makeatletter
\newcommand*{\rom}[1]{\expandafter\@slowromancap\romannumeral #1@}
\makeatother

\usepackage{hyperref}

\usepackage[accepted]{icml2022}

\icmltitlerunning{Generalization, Mayhems and Limits in Recurrent Proximal Policy Optimization}

\begin{document}

\twocolumn[
\icmltitle{Generalization, Mayhems and Limits in Recurrent Proximal Policy Optimization}




\begin{icmlauthorlist}
\icmlauthor{Marco Pleines}{tu}
\icmlauthor{Matthias Pallasch}{tu}
\icmlauthor{Frank Zimmer}{hsrw}
\icmlauthor{Mike Preuss}{nl}
\end{icmlauthorlist}

\icmlaffiliation{tu}{Department of Computer Science, TU Dortmund University, Germany}
\icmlaffiliation{hsrw}{Faculty of Communication and Environment, Hochschule Rhein-Waal, Kamp-Lintfort, Germany}
\icmlaffiliation{nl}{LIACS, Universiteit Leiden, Leiden, Netherlands}


\icmlkeywords{Deep Reinforcement Learning, Proximal Policy Optimization, Truncated Backpropagation Through Time, Recurrence, GRU, LSTM, Recurrent Policy Gradient}

\vskip 0.3in
]



\printAffiliationsAndNotice{}  

\begin{abstract}
At first sight it may seem straightforward to use recurrent layers in Deep Reinforcement Learning algorithms to enable agents to make use of memory in the setting of partially observable environments.
Starting from widely used Proximal Policy Optimization (PPO), we highlight vital details that one must get right when adding recurrence to achieve a correct and efficient implementation, namely: properly shaping the neural net's forward pass, arranging the training data, correspondingly selecting hidden states for sequence beginnings and masking paddings for loss computation.
We further explore the limitations of recurrent PPO by benchmarking the contributed novel environments Mortar Mayhem and Searing Spotlights that challenge the agent's memory beyond solely capacity and distraction tasks.
Remarkably, we can demonstrate a transition to strong generalization in Mortar Mayhem when scaling the number of training seeds, while the agent does not succeed on Searing Spotlights, which seems to be a tough challenge for memory-based agents.
\end{abstract}
\section{Introduction}

Leveraging recurrent neural networks in Deep Reinforcement Learning (DRL) enabled agents to possess memory that can provide information on the history of the current task that is not present in the agent's contemporary observation.
Exploiting such memory is necessary to solve many Partially Observable Markov Decision Processes (POMDP).
This led to many emergent and compelling agent behaviors that play complex games like Capture the flag \cite{Jaderberg2018}, StarCraft \rom{2} \cite{Vinyals2019Starcraft2}, DotA 2 \cite{Berner2019Dota2} and Hide and Seek \cite{Baker2020HideAndSeek}.
The utilization of recurrent layers, such as the widely used long short-term memory (LSTM) \cite{Hochreiter1997} and gated recurrent unit (GRU) \cite{Cho2014GRU}, notably increases the complexity and thus the error-proneness of DRL algorithms.

\begin{figure}[t]
\vskip 0.2in
\begin{center}
\centerline{\includegraphics[width=\columnwidth]{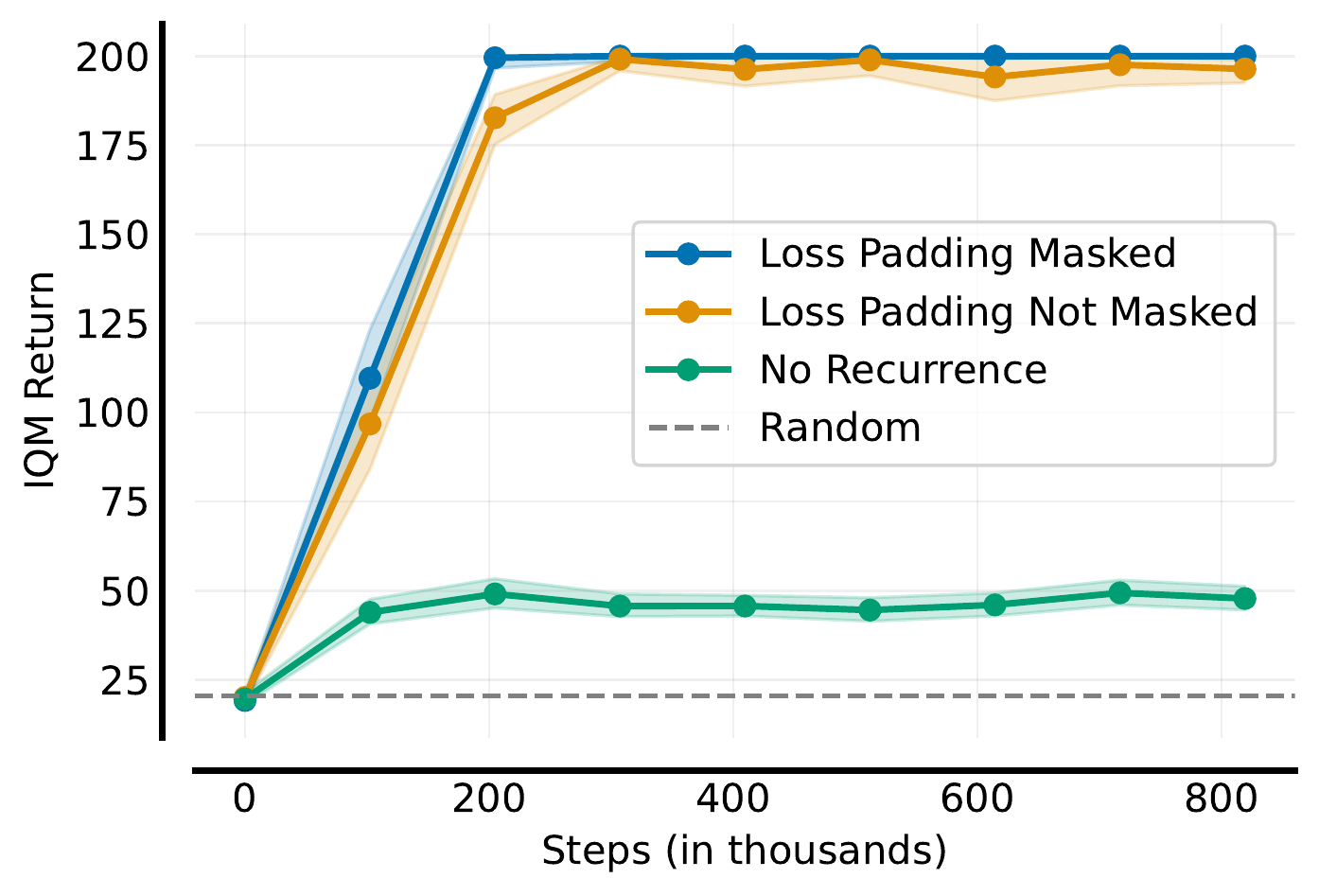}}
\caption{CartPole can be turned into a POMDP by removing the velocity from the observation space. For instance, if paddings are not masked during loss computation, somewhat good results might mislead to false conclusions. Each experiment was trained 5 times and evaluated on 10 novel seeds 3 times.}
\label{fig:cartpole}
\end{center}
\vskip -0.3in
\end{figure}

Adding recurrence to DRL algorithms, as PPO, seems to be largely considered straightforward, which leaves a lot of details to engineering that however plays a vital role in obtaining a correctly working and efficient implementation\footnote{Bugs that hindered progress are described in appendix \ref{sec:annex_bugs}.}.
We may consider four major issues that must be resolved:

\begin{compactitem}
    \item efficiently organizing the model's forward pass
    \item data handling concerning unequal sequence lengths
    \item correctly sampling hidden states for sequence starts
    \item mask paddings while computing the loss
\end{compactitem}


Correctly setting up the way training data is structured, propagated through the neural net and used for optimization is quite an effort and prone to errors.
There is also the risk of misleading results on low-scale problems to verify an implementation's correctness, as portrayed by figure \ref{fig:cartpole}.
It may be for this reason that how to do it has not been thoroughly documented in literature according to our knowledge.
This shortcoming entails possible differences in research works that may actually lead to quite diverse results.

We believe this work is the first to precisely describe how to successfully add a recurrent layer to the on-policy gradient algorithm Proximal Policy Optimization (PPO) \cite{Schulman2017}, focusing especially on pixel input (visual observation) and discrete action spaces.
Additionally, a densely commented and documented baseline implementation\footnote{\url{https://github.com/MarcoMeter/recurrent-ppo-truncated-bptt}}, supporting truncated backpropagation through time, is contributed.


Next to these algorithmic adaptations, we also contribute two novel environments challenging the agent's memory, called Mortar Mayhem and Searing Spotlights.
These environments
enrich the landscape of assessing memory-based agents, because these are not solely limited to capacity and distraction problems, as most well-established environments in this field.
Examining the capacity and the robustness to distractions is not unique to DRL as it can be applied to a much broader context employing a recurrent, transformer or similar architecture.
Both environments require the agent not just to maintain initial goal cues in its memory for many steps, but also to continuously modify its memory to memorize its past behavior to solve the task at hand.

We show that Mortar Mayhem's second task cannot be trained without memory.
Strong generalization is presented on its complete task, which is notably related to scaling the number of training seeds. 
However, this effect does not apply to the ballet environment, leaving us with weak generalization as already shown by \citeauthor{Lampinen2021htm} \yrcite{Lampinen2021htm}.
Searing Spotlights pushes recurrent PPO to its limits.
Only simplified versions of this environment are solvable to some extent.
Further, refreshing stale hidden states and advantages does not provide a gain in performance that satisfies the gain in computational cost.


\section{Related Work}

\subsection{Proximal Policy Optimization and Recurrence}

Revisiting literature on PPO helps to understand significant design and implementation decisions \cite{Henderson2018, Engstrom2020Implementation, Chloe2020, andrychowicz2021what}.
Related work, in the domain of PPO using recurrence on control tasks (MountainCar, Ant, HalfCheetha, ect.) with continuous action spaces and state-based inputs (i.e. vector observation), examines generalization \cite{Packer2018}, meta reinforcement learning \cite{Yu2019MetaRL} and robust reinforcement learning \cite{Zhang2021robustRL}.
None of the aforementioned publications describe the specifics of getting recurrent PPO to work, which is actually a non-trivial process.
In contrast to those, all our experiments are run on environments that feature discrete action spaces and pixel input.
Our work also details all the necessary components to hook up a recurrent layer.

\subsection{Applications of Recurrent Layers in DRL}

Next to PPO, other algorithms were published that employ recurrence.
\citeauthor{Hausknecht2015DRQN} \yrcite{Hausknecht2015DRQN} extended the Deep Q-Network (DQN) algorithm and examined the way how episodes and their hidden states are sampled from the experience replay buffer.
More literature is available on adding recurrence to variants of Deep Deterministic Policy Gradient (DDPG) algorithms \cite{Heess2015rdpg, Meng2021TD3}.
Especially \citeauthor{Ni2021} \yrcite{Ni2021} proposed a recurrent Twin-Delayed DDPG implementation and shared how their replay buffer is made more memory-efficient.
The authors of the off-policy algorithm R2D2 (Recurrent Replay Distributed DQN) provide approaches to sample sequences from the replay buffer, to initialize hidden states and to mitigate their staleness \cite{kapturowski2019r2d2}.
\citeauthor{Kamarl2020} \yrcite{Kamarl2020} showed potential to mitigate such staleness in PPO by refreshing the hidden states every few minibatch updates.
Contrary to their findings, our experiments show no notable improvement when refreshing hidden states during optimization.

\subsection{Partially Observable Environments as Benchmarks}

Typical POMDPs for assessing recurrent DRL algorithms are the Morris water maze \cite{DHooge2001watermaze} and the T-maze \cite{Wiestra2007}.
The T-maze asks the agent to initially memorize a goal cue, then to traverse a long alley and once the end is reached it has to choose the correct exit.
Inside the water maze, the agent has to find an invisible goal platform and memorize its location.
Once found, the agent is set to a random position and has to find the goal again.
The Memory Task Suite \cite{fortunato2019generalization} comprises 13 tasks to assess memory in the broader context of machine learning.
Some variants build on top of the idea of the water maze (Goal Navigation), whereas others extend the mechanics of the T-maze (Spot the Difference).
A few environments, like dancing the ballet, are contributed by \citeauthor{Lampinen2021htm} \yrcite{Lampinen2021htm}.
Control tasks are normally fully observable.
These are turned into POMDPs by adding noise to the observations or by removing information, such as the velocity, from the observation space.
Atari games are usually trained using a frame stack of past observations.
Removing that stack and adding flickering frames transforms these tasks to POMDPs as well \cite{Hausknecht2015DRQN}.
Our work contributes two novel environments providing unique challenges to memory-based DRL agents.

\begin{figure*}
    \centering
    \includegraphics[width=\textwidth*\real{0.95}]{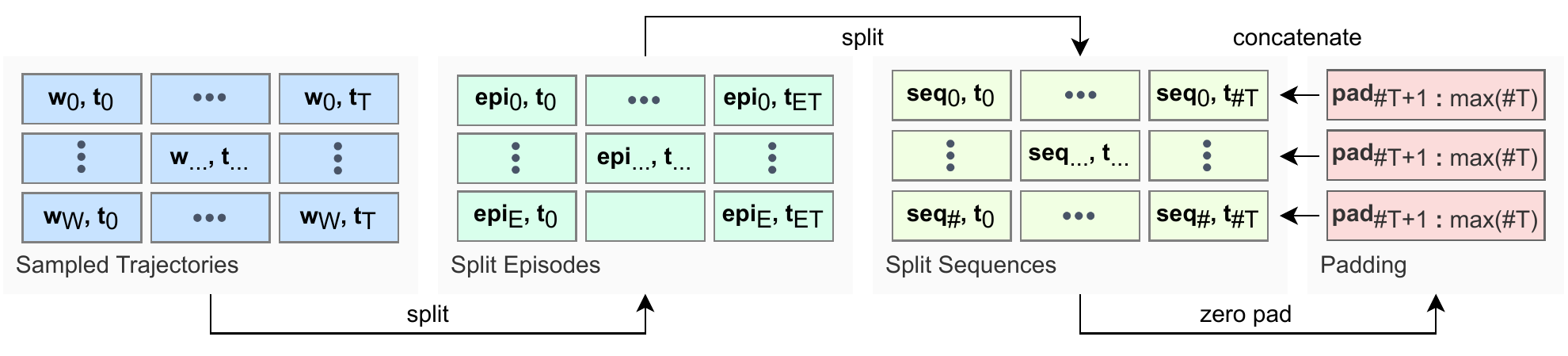}
    \caption{The data preprocessing starts out by sampling trajectories across $W$ workers for $T$ steps. Next, $E$ episodes of varying length $ET$ are extracted from the trajectories.
    Those can be further split into $\#$ sequences of varying length $\#T$.
    At last, zero padding is used to retrieve sequences of fixed length $max(\#T)$.}
    \label{fig:data_prep}
\end{figure*}

\section{Recurrent Proximal Policy Optimization}

The agent is trained using Proximal Policy Optimization (PPO) using the clipped surrogate objective \cite{Schulman2017}.
Due to the usage of a recurrent layer (e.g. LSTM), the to be selected action $a_t$ of the policy $\pi_\theta$ depends on the current observation $o_t$ and hidden state $h_t$ of the recurrent layer.
$\hat{A}_t$ denotes advantage estimates based on generalized advantage estimation (GAE), $\theta$ the parameters of a neural net and $\epsilon$ the clip range.

\small
\begin{equation}
    L^{C}_t(\theta) = \hat{\mathbb{E}}_t [min(q_t(\theta)\hat{A}_t,clip(q_t(\theta),1- \epsilon,1+\epsilon)\hat{A}_t)]
\end{equation}
\begin{equation*}
    \textnormal{with ratio}~q_t(\theta) = \frac{\pi _\theta(a_t|o_t,h_t)}{\pi _{\theta \text{old}}(a_t|o_t, h_t)}
\end{equation*}
\normalsize

The value function is optimized using the squared-error loss $L_t^V(\theta)$.
$\mathcal{H}[\pi_\theta](o_t)$ denotes an entropy bonus encouraging exploration \cite{Schulman2017}.
Both are weighted using the coefficients $c_1$ and $c_2$ and are added to $L_t^C$ to complete the loss:

\small
\begin{equation}
	L^{CVH}_t(\theta)=\hat{\mathbb{E}}_t [L^{C}_t(\theta)-c_1L^{V}_t(\theta)+c_2\mathcal{H}[\pi_\theta](o_t, h_t)]
\end{equation}
\begin{equation*}
    \textnormal{with}~ L^{V}_t(\theta)=(V_\theta(o_t, h_t)-V_t^{targ})^2
\end{equation*}
\normalsize

\begin{figure*}
	\centering
	\includegraphics[width=0.85\textwidth]{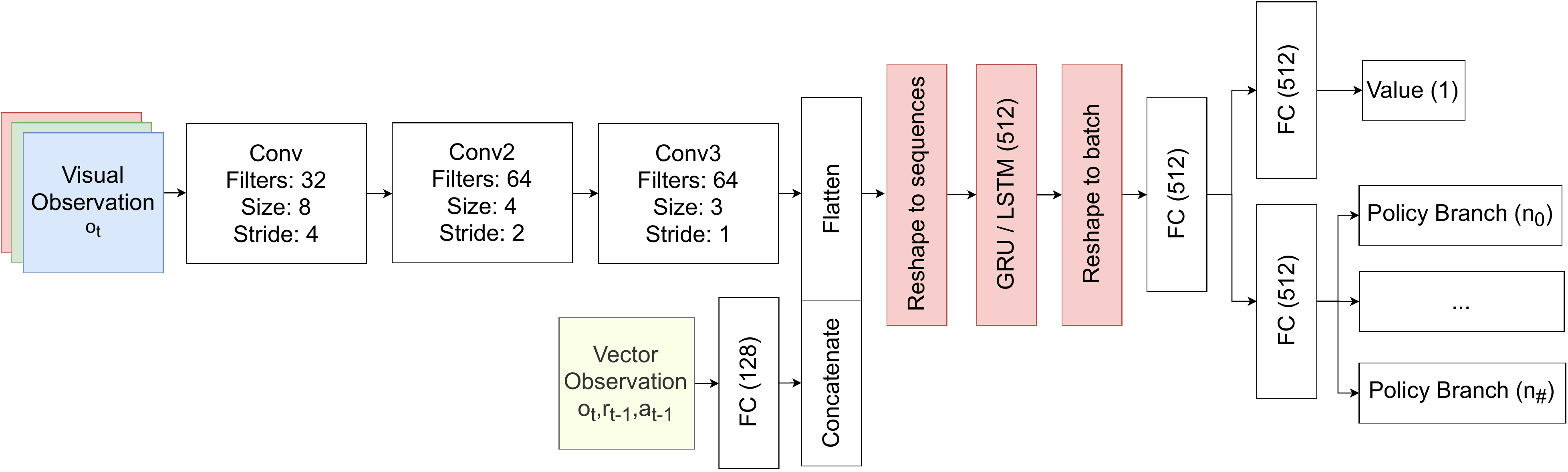}
	\caption[Model]{Recurrent PPO utilizes a feed-forward convolutional recurrent neural network.
	First, visual and vector observations are encoded as an entire batch by either convolutional or fully connected layers.
	$o_t$, $a_{t-1}$ and $r_{t-1}$ denote the current observation, the last action and the last reward.
	The encoded features are concatenated and then reshaped to sequences before feeding them to the recurrent layer.
	Its output has to be reshaped into the original batch shape.
	Further, the forward pass is divided into two streams relating to the value function and the policy.
	The number of policy heads is equal to the number of action dimensions given by a multi-discrete action space.}
	\label{fig:model}
	\vskip -0.1in
\end{figure*}

\subsection{Fundamental Implementation Details}

Adding a recurrent layer to a model that is trained with PPO is not just plug-and-play.
To ensure an efficient and correct implementation, three major components have to be touched: the processing of the sampled training data (Figure \ref{fig:data_prep}), the forward pass of the model (Figure \ref{fig:model}) and the loss function (Equation \ref{eq:mask}).

The training data is sampled by a fixed number of workers for a fixed amount of steps and is stored as a tensor.
Each collected trajectory may contain multiple episodes that might have been truncated.
After sampling, the data has to be split into episodes.
Optionally, these can be further split into smaller sequences of fixed length.
Otherwise, the actual sequence length is equal to the length of the longest episode.
Episodes or sequences that are shorter than the desired length are padded using zeros.
As the data is structured into fragments of episodes, the hidden states of the recurrent layer have to be selected correctly.
The output hidden state of the previous sequence is the input hidden state of its consecutive one.
This approach is also known as truncated backpropagation through time (truncated bptt).
Finally, minibatches sample multiple sequences from the just processed data.

Concerning the forward pass of the model, it is more efficient to feed every non-recurrent layer the entire batch of the data (i.e. $batch~size = workers \times steps$)  and not each sequence one by one.
Whenever the batch is going to be fed to a recurrent layer during optimization, the batch has to be reshaped to the dimensions: number of sequences and sequence length.
After passing the sequences to the recurrent layer, the data has to be reshaped again to the overall batch size.
Note that the forward pass for sampling trajectories operates on a sequence length of one.
In this case, the data keeps its shape throughout the entire forward pass.

Once the loss function is being computed, the padded values of the sequences have to be masked so that these do not affect the gradients.
$L^{mask}$ is the average over all losses not affected by the paddings.



\begin{small}
\begin{equation}
    L^{mask}(\theta)= \frac{\sum_t^T \left [ mask_t \times L^{CVH}_t(\theta) \right ]}{\sum_t^T[mask_t]}
    \label{eq:mask}
\end{equation}

\begin{equation*}
    \textnormal{with}~mask_t=
    \begin{cases}
        0 & \text{where padding is used} \\
        1 & \text{where no padding is used} \\
    \end{cases}
\end{equation*}
\end{small}



\subsection{Refreshing Stale Advantages and Hidden States}

    
Once the parameters of the model are updated using at least one minibatch, the consecutive minibatch updates of the current iteration operate to some extent on stale data.
\citeauthor{andrychowicz2021what} \yrcite{andrychowicz2021what} empirically identified that recomputing advantage estimates before each epoch can be beneficial to policy gradient algorithms given the on-policy setting.
This measure can also be applied to hidden states of a recurrent layer, as recurrent staleness was also observed by \citeauthor{kapturowski2019r2d2} \citeyear{kapturowski2019r2d2}.
According to our knowledge, refreshing stale hidden states were only briefly examined by \citeauthor{Kamarl2020} \yrcite{Kamarl2020} on the on-policy algorithm PPO.
The contributed recurrent PPO algorithm encompasses the possibility to refresh advantages and hidden states before each epoch to mitigate their staleness.
Finally, algorithm \ref{alg:ppo} describes our realization of recurrent PPO.

\begin{algorithm}[h]
   \caption{Recurrent PPO using Truncated BPTT}
   \label{alg:ppo}
\begin{algorithmic}
   \FOR{iteration=$1,2,\ldots$}
   \FOR{worker=$1,2,\ldots,W$}
        \STATE Run policy $\pi_{\theta old}$ in environment for $T$ timesteps
   \ENDFOR
   \STATE Compute advantage estimates $\hat{A}_{1,1},\ldots,\hat{A}_{T, W}$
   \STATE Split trajectories into episodes
   \IF{fixed sequence length}
    \STATE Split episodes into sequences of fixed length
    \ELSE
    \STATE sequences = episodes
   \ENDIF
   \STATE Zero pad sequences that are too short
   \STATE Select initial hidden states of each sequence
   \FOR{epoch}
        \FOR{minibatch=$1,2,\ldots MB$ with size $M \leq WT$}
        \STATE Optimize $L^{mask}(\theta)$
        \ENDFOR
        \IF{epoch $> 0$}
        \STATE Recompute hidden states and advantages w.r.t. $\theta$
        \ENDIF
        \STATE $\theta_{old} \leftarrow \theta$
   \ENDFOR
   \ENDFOR
\end{algorithmic}
\end{algorithm}
\section{Novel Environments}
\hyphenation{MiniGrid}

Many partially observable environments exist that challenge the agent's memory.
The majority of them, such as MiniGrid Memory \cite{Chevalier-Boisvert2018minigrid}, Spot the Difference \cite{fortunato2019generalization} and Remembering the Ballet \cite{Lampinen2021htm} demand the agent's memory to solely memorize the to be observed goal cues during the very beginning of the episode.
Once this cue is memorized, there is no need to further manipulate the agent's memory.
Simply maintaining it is enough to solve the task.
It can be hypothesized that the extracted features from observing the goal cues are sufficient to solve the ballet environment.
To show this, the ballet environment can be made fully observable (markovization) by feeding the entire sequence of cues to a recurrent encoder that extracts features, which are utilized by the agent's policy.
Consequently, the policy itself does not need to maintain a memory in this case.
The task at hand gets much easier, because the agent does not need to make obsolete decisions while observing the sequence of cues as its position is frozen.
To investigate a higher level of complexity, the environments Mortar Mayhem and Searing Spotlights are introduced next.
These require the agent to properly manipulate its memory throughout an entire episode, because it will be relevant to the agent to memorize not just only initial observations (e.g. cues), but especially also the past decisions that it has made.
All possible environment configurations and exemplary episodes can be found in Appendix \ref{sec:annex_mortar} and \ref{sec:annex_searing}.

\begin{figure}
\begin{center}
\subfigure[Clue Task]{\label{fig:a}\includegraphics[width=\columnwidth*\real{0.325}]{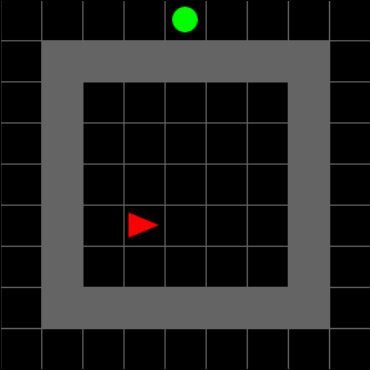}}
\subfigure[Act Task]{\label{fig:b}\includegraphics[width=\columnwidth*\real{0.325}]{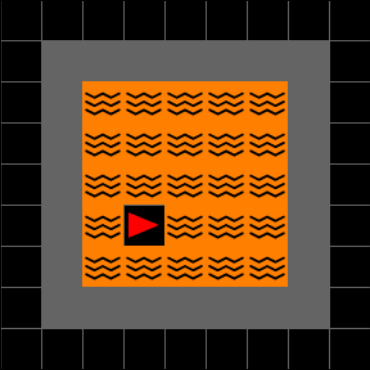}}
\subfigure[Act Only Task]{\label{fig:b_only}\includegraphics[width=\columnwidth*\real{0.325}]{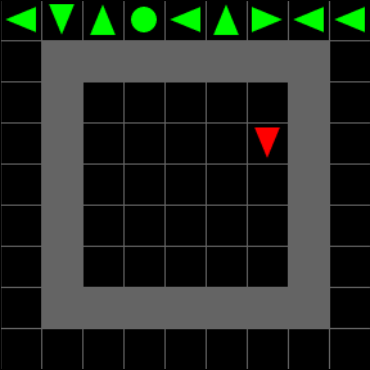}}
\caption{In Mortar Mayhem, the agent has to memorize a sequence of commands first (a) and then has to execute them one by one. Between commands, the agent has 5 steps to move to the commanded tile. Once the time is up, lava tiles (b) verify the correct position of the agent. The environment can also be run given all commands at once during each step (c).}
\label{fig:mm}
\end{center}
\vskip -0.3in
\end{figure}

\subsection{Mortar Mayhem}

The Mortar Mayhem environment (Fig. \ref{fig:mm}) is inspired by the commercial video game Pummel Party\footnote{\url{http://rebuiltgames.com/}} and is implemented using the minimalistic gridworld environment \cite{Chevalier-Boisvert2018minigrid}.
Two tasks make up the dynamics of the environment.
At first, the agent has to memorize a sequence of commands (Clue Task) and afterwards it has to execute each command in the observed order (Act Only Task).
One command orders the agent to move to one adjacent floor tile or to stay at the current one.
If the agent fails, the episode terminates, while receiving no reward.
Upon successfully executing one command, a reward of $+0.1$ is signaled.
Mortar Mayhem's observation space comprises 490 values that are retrieved by encoding the underlying grid world one-hot.
Alternatively, the observation space can be represented by $84\times84\times3$ pixels.
The action space is discrete and provides the actions move forward, rotate left, rotate right and stay.
In contrast to environments as the aforementioned ones, the agent has to continuously modify its memory to solve the second task.
It has to memorize which orders were already taken to succeed on the consecutive command.
    
\begin{figure}
\centering
\subfigure[Initial state]{\label{fig:s_a}\includegraphics[width=\columnwidth*\real{0.47}]{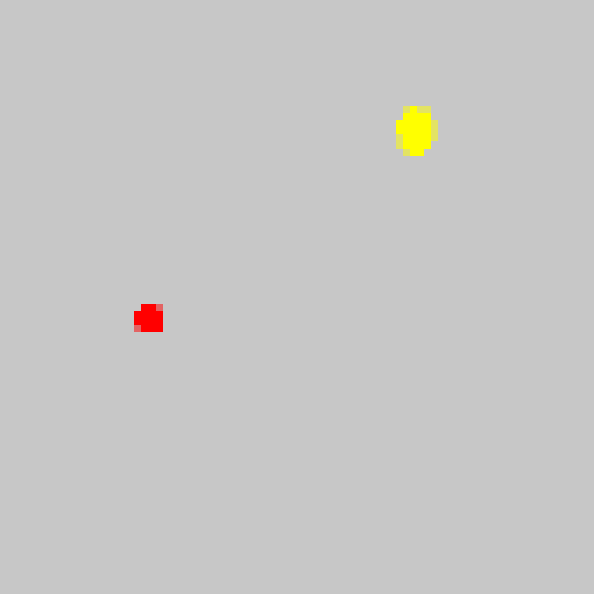}}
\subfigure[In progress]{\label{fig:s_b}\includegraphics[width=\columnwidth*\real{0.47}]{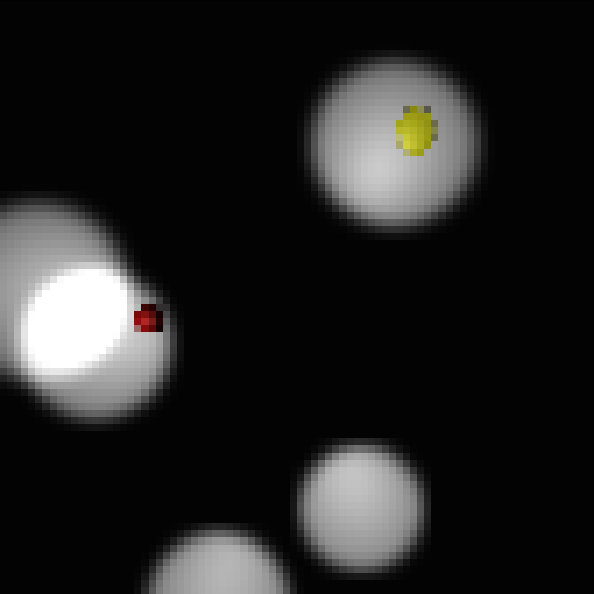}}
\caption{In Searing Spotlights, the environment is fully observable for a few initial steps (a) and then dims the light until off. The agent is the red circle, while coins are illustrated by yellow ellipses. Over time, more and more harmful spotlights move across the environment (b). The agent has to collect all coins, while not being unveiled by a spotlight.}
\label{fig:ss}
\vskip -0.15in
\end{figure}

\subsection{Searing Spotlights}

Searing Spotlights (Figure \ref{fig:ss}) is inspired by Pummel Party as well and is implemented using the ml-agents toolkit \cite{Juliani2018mlagents}.
The agent's primary goal in this environment is to avoid being unveiled by spotlights.
Initially, fewer spotlights traverse the environment, but over time these are more frequently spawned.
Thus, the environment is less hostile early on, while providing little information.
As time progresses, more information is available to the agent, but the environment is more dangerous.
After all, the agent is especially challenged to memorize its past decisions to estimate its current location.
To avoid numb policies, a coin collection task is added to the environment.
This ensures that the agent has to move across the environment and therefore putting itself more at risk.
Searing spotlights features an input of $84\times84\times3$ pixels.
The agent's action space is multi-discrete comprising two discrete dimensions.
Moving left, right and not moving horizontally are part of the first dimension.
The second dimension enables the agent to move up, down or not move vertically at all.
Concerning the reward function, the agent is punished for being visible inside a spotlight by a reward of $-0.01$ for each step, while it is being rewarded by $+1.0$ for collecting a coin.
Terminal states are reached once the agent loses all of his 100 hit points (spotlights damage the agent by $0.01$ every step) or once all coins are collected.

\begin{table}
\caption{Parameters used for training.
Vector observations combine the current observation, the last action (one-hot) and reward.
The learning rate $\alpha$ and the entropy bonus coefficient $c_2$ decay linearly.
The parameters on the bottom part are the same for each environment.}
\vskip 0.1in
\begin{small}
\centering
\begin{tabular}{l|rrr}
\rowcolor[HTML]{EFEFEF} 
\textbf{Parameter}  & \multicolumn{1}{c}{\cellcolor[HTML]{EFEFEF}\textbf{Mortar}} & \multicolumn{1}{c}{\cellcolor[HTML]{EFEFEF}\textbf{Searing}} & \multicolumn{1}{c}{\cellcolor[HTML]{EFEFEF}\textbf{Ballet}} \\ \hline
Seeds               & 1000                                                        & 1000                                                         & 500                                                         \\
\rowcolor[HTML]{EFEFEF} 
Visual Observation  & 84$\times$84$\times$3                                                     & 84$\times$84$\times$3                                                      & 99$\times$99$\times$3                                                     \\
Vector Observation  & 490+4+1                                                     & 6+1                                                          & 13+8+1                                                      \\
\rowcolor[HTML]{EFEFEF} 
Action Space        & 4                                                           & 3, 3                                                         & 8                                                           \\
Vector Encoder Size & 256                                                         & 128                                                          & 128                                                         \\
\rowcolor[HTML]{EFEFEF} 
Number of Workers   & 32                                                          & 16                                                           & 32                                                          \\
Batch Size          & 16384                                                       & 8192                                                         & 16384                                                       \\
\rowcolor[HTML]{EFEFEF} 
Epochs              & 4                                                           & 3                                                            & 3                                                           \\
Initial $\alpha$    & 0.0003                                                      & 0.0003                                                       & 0.0003                                                      \\
\rowcolor[HTML]{EFEFEF} 
Min $\alpha$        & 0.00003                                                     & 0.0003                                                       & 0.0003                                                      \\
Initial $c_2$       & 0.1                                                         & 0.0001                                                       & 0.001                                                       \\
\rowcolor[HTML]{EFEFEF} 
Min $c_2$           & 0.0001                                                      & 0.0001                                                       & 0.0005                                                      \\
Sequence Length     & Max                                                         & 128                                                          & Max                                                         \\
\rowcolor[HTML]{EFEFEF} 
Discount Factor     & \multicolumn{3}{c}{\cellcolor[HTML]{EFEFEF}0.99}                                                                                                                                         \\
Lamda (GAE)         & \multicolumn{3}{c}{0.95}                                                                                                                                                                 \\
\rowcolor[HTML]{EFEFEF} 
Worker Steps        & \multicolumn{3}{c}{\cellcolor[HTML]{EFEFEF}512}                                                                                                                                          \\
Minibatches         & \multicolumn{3}{c}{8}                                                                                                                                                                    \\
\rowcolor[HTML]{EFEFEF} 
$c_1$               & \multicolumn{3}{c}{\cellcolor[HTML]{EFEFEF}0.25}                                                                                                                                         \\
Max Gradient Norm   & \multicolumn{3}{c}{0.5}                                                                                                                                                                  \\
\rowcolor[HTML]{EFEFEF} 
Clip Range          & \multicolumn{3}{c}{\cellcolor[HTML]{EFEFEF}0.2}                                                                                                                                          \\
Hidden State Size   & \multicolumn{3}{c}{512}                                                                                                                                                                  \\
\rowcolor[HTML]{EFEFEF} 
Recurrent Layer     & \multicolumn{3}{c}{\cellcolor[HTML]{EFEFEF}GRU}                                                                                                                                          \\
Activations         & \multicolumn{3}{c}{ReLU}                                                                                                                                                                 \\
\rowcolor[HTML]{EFEFEF} 
Optimizer           & \multicolumn{3}{c}{\cellcolor[HTML]{EFEFEF}AdamW}                                                                                          
\end{tabular}
\label{tab:config_complete}
\end{small}
\vskip -0.1in
\end{table}
\section{Experiments}

Every provided sample efficiency curve shows the interquartile mean (IQM) and confidence interval (95\%) as recommended by the evaluation library \textit{rliable} \cite{agarwal2021statistics}.
Each experiment trained 5 agents.
The result data is retrieved from evaluating 10 training and 10 novel environment seeds, which were repeated 3 times.
Thus, one data point aggregates 150 episodes with respect to training and novel seeds.
Table \ref{tab:config_complete} enumerates the used training parameters.
Based on grid search (Appendix \ref{sec:grid_search}), the hidden state size, the sequence length and the recurrent layer type were selected.
GRU turned out to be slightly more effective than LSTM in the trained environments.
The source code is available online\footnote{\url{https://github.com/MarcoMeter/neroRL}}.

\subsection{Act Only Task in Mortar Mayhem}

\begin{figure}
\begin{center}
\centerline{\includegraphics[width=\columnwidth]{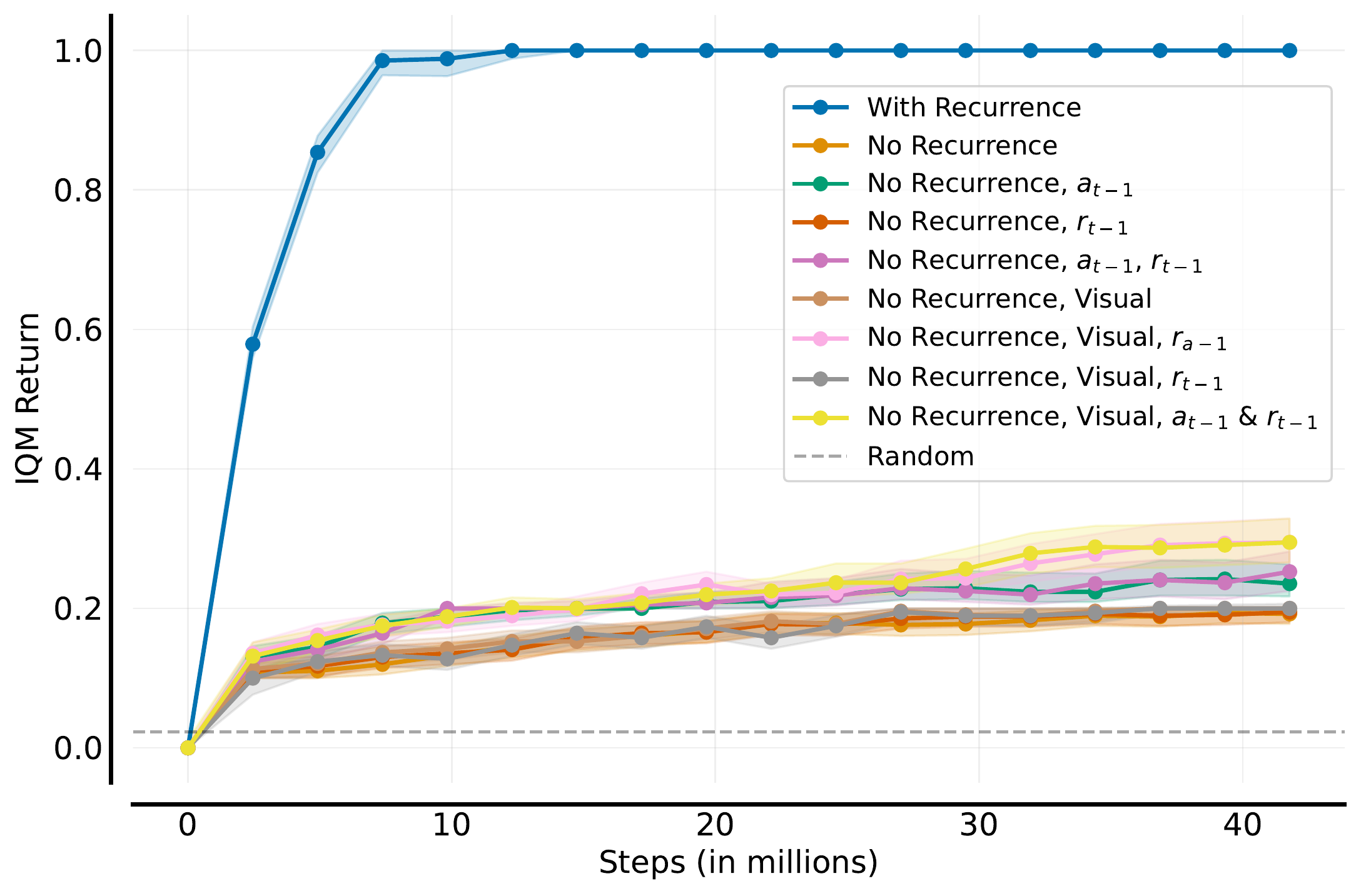}}
\caption{Training performance of the agent with and without recurrence on the Act Only Task in Mortar Mayhem.}
\label{fig:mortar_b_train}
\end{center}
\vskip -0.2in
\end{figure}

To verify the second task of Mortar Mayhem, the agent is asked to complete 10 commands with and without recurrence, while all commands are visible during each step throughout one episode.
Figure \ref{fig:mortar_b_train} presents the training performance where the observation space of the agent is varied by providing several permutations of using visual or vector observations and whether the agent may perceive its last action and last reward.
Recurrence is essential to solve the Act Task, while visual observations are slightly better than their vector counterparts if no memory is present.
Under these circumstances, the agent slightly profits from sensing its last action, but this small impact is not observed for its last reward.
It can be argued that more model parameters (e.g. due to the convolutional layers) and the info of the last action aid overfitting.
None of these training runs surpassed an IQM return of $0.1$ given novel seeds.

\begin{figure}[H]
\begin{center}
\centerline{\includegraphics[width=\columnwidth]{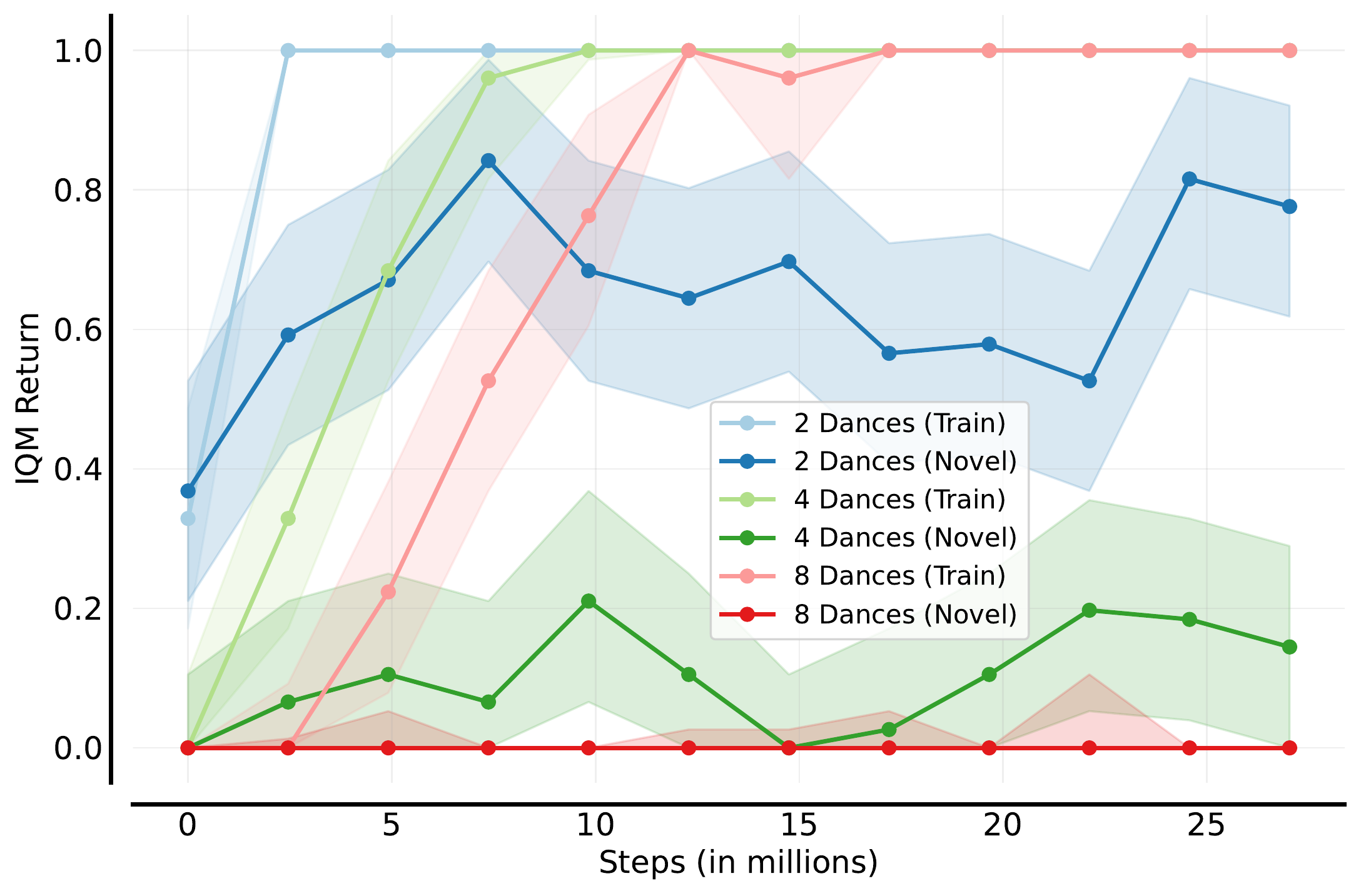}}
\caption{Performance on novel and training seeds in the ballet environment assessed on either 2, 4, or 8 dances.}
\label{fig:ballet_selected}
\end{center}
\vskip -0.4in
\end{figure}

\subsection{Generalization}

\begin{figure}
\begin{center}
\centerline{\includegraphics[width=\columnwidth]{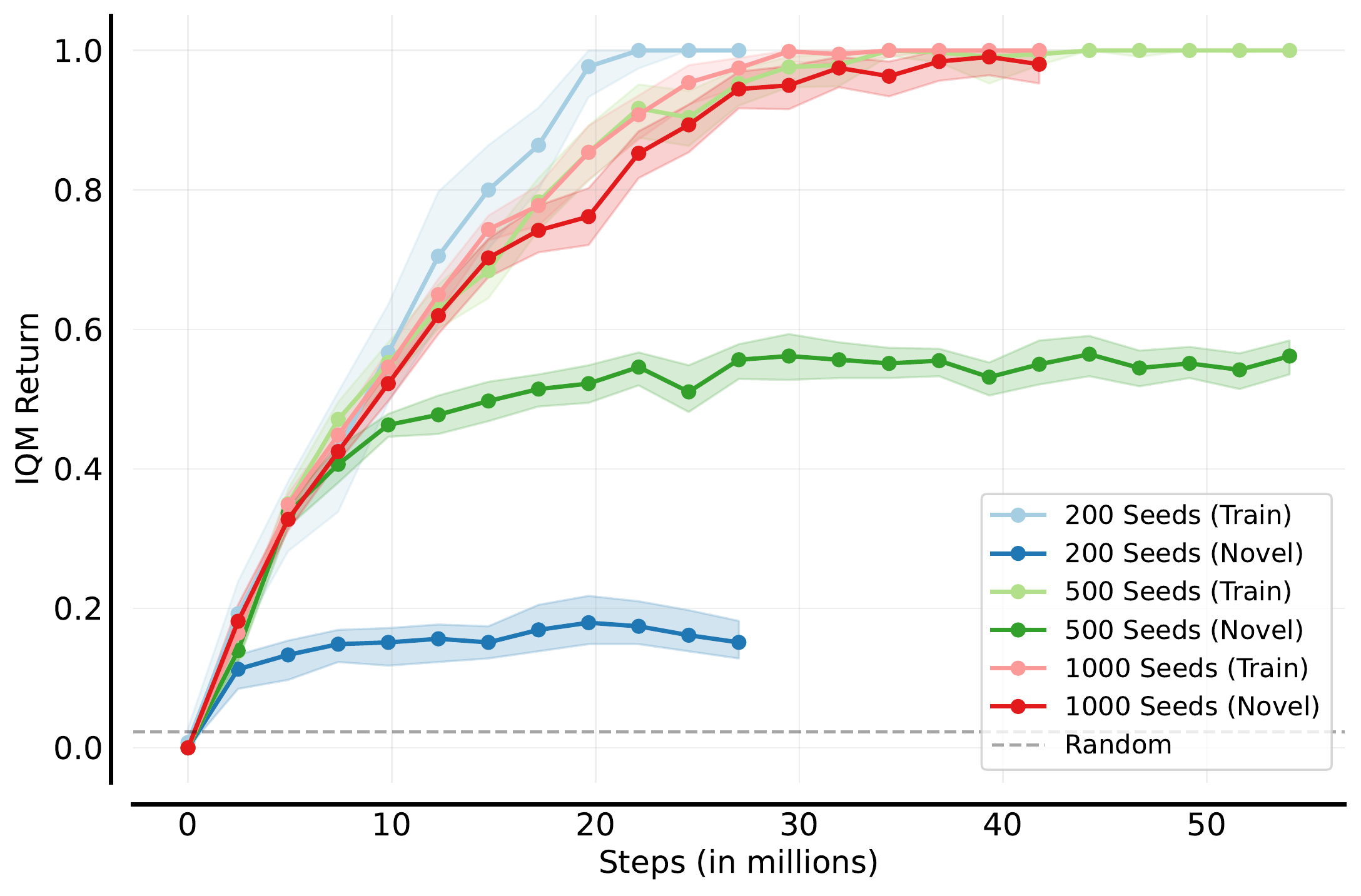}}
\caption{Generalization performance on training either 200, 500 or 1000 seeds in Mortar Mayhem.}
\label{fig:mortar_seeds}
\end{center}
\vskip -0.2in
\end{figure}

Generalization is examined on Mortar Mayhem and the ballet environment.
The agent generalizes well in the former one when trained on 10 commands (Figure \ref{fig:mortar_seeds}).
However, 1000 seeds are required to achieve this performance, while 500 only reach an IQM return of less than $0.6$.
The scale of 200 seeds is by far inferior.
Concerning the ballet environment (Figure \ref{fig:ballet_selected}), the agent has no problems solving training seeds, while its performance on novel seeds may be worse than chance\footnote{Chance depends on the number of dances.}.
Our results pretty much confirm the ones reported by \citeauthor{Lampinen2021htm} \yrcite{Lampinen2021htm}.
However, there are a few apparent differences in the training setup.
We trained the agent only for about 25 million steps (not 5 billion).
The dance to be identified is one-hot encoded as part of the vector observation space, whereas \citeauthor{Lampinen2021htm} \yrcite{Lampinen2021htm} use a recurrent language encoder.
During training, the number of dances varied between 2, 4 and 8 dances, while the delay between dances was held constant at 16.
Scaling up the number of training seeds to 1000 does not yield any improvement over 500 ones.

\subsection{Refreshing Stale Hidden States and Advantages}
\label{sec:refresh}

\begin{figure}[t]
\begin{center}
\centerline{\includegraphics[width=\columnwidth]{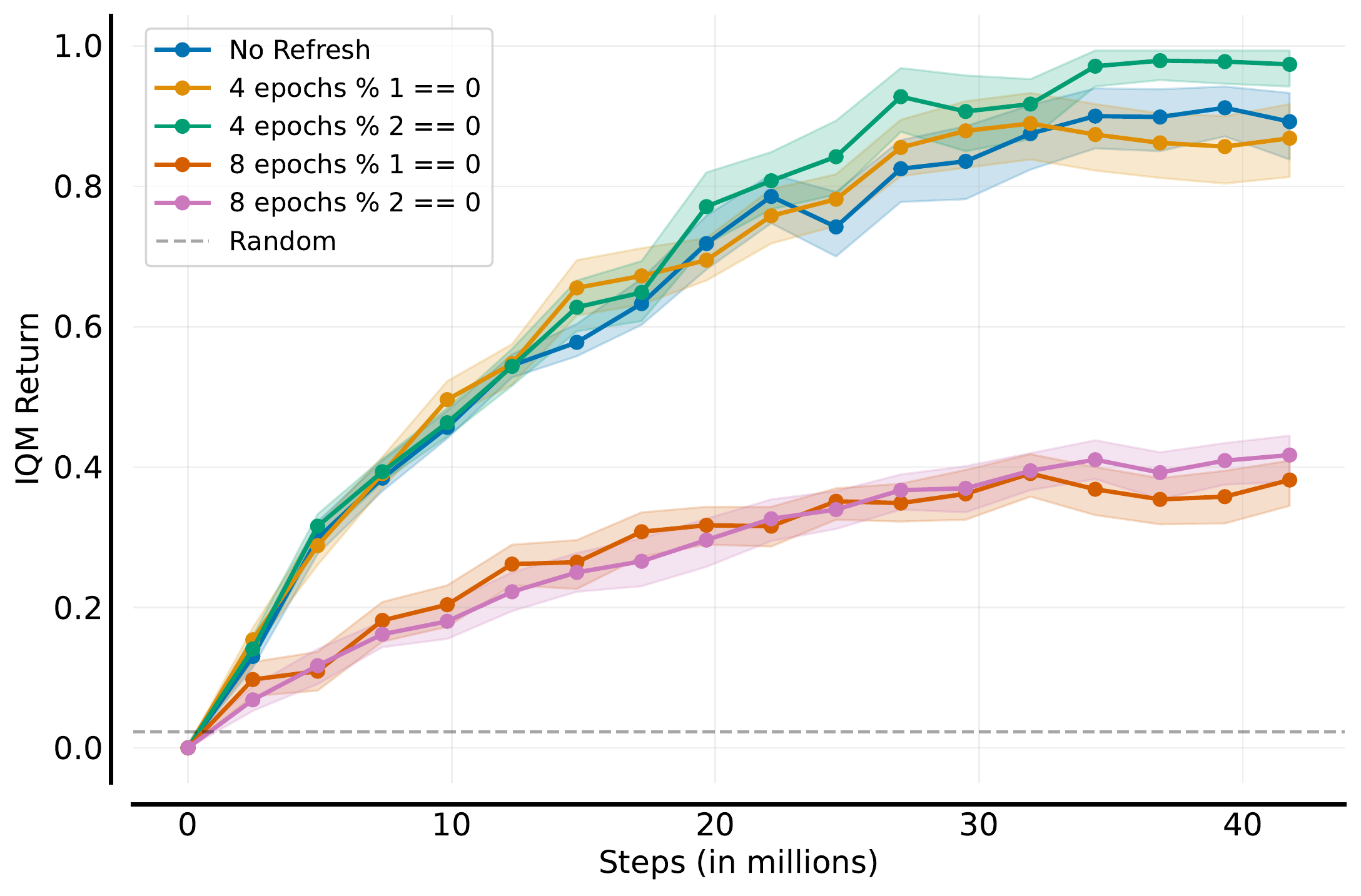}}
\caption{Generalization performance on novel seeds when stale hidden states and advantages are refreshed in Mortar Mayhem.}
\label{fig:mortar_refresh_novel}
\end{center}
\vskip -0.2in
\end{figure}

\begin{figure}[t]
\begin{center}
\centerline{\includegraphics[width=\columnwidth]{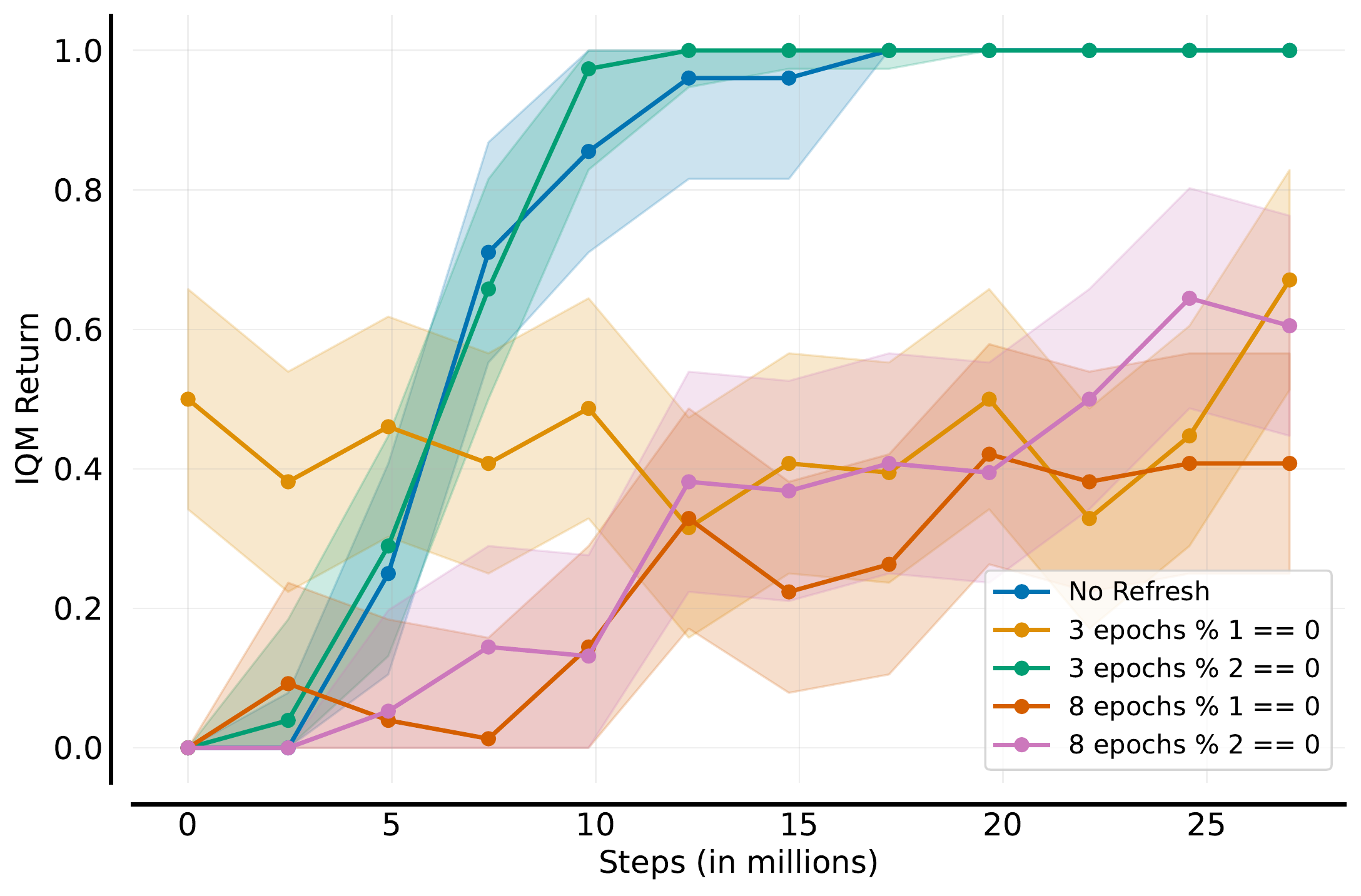}}
\caption{Training performance on 8 dances when stale hidden states and advantages are refreshed in the ballet environment.}
\label{fig:ballet_refresh_train}
\end{center}
\vskip -0.2in
\end{figure}

Refreshing stale hidden states and advantages does not yield promising results.
Applying the refresh every other epoch slightly improves the performance in Mortar Mayhem on novel seeds (Figure \ref{fig:mortar_refresh_novel}) and ballet on training seeds (Figure \ref{fig:ballet_refresh_train}).
This minor improvement is likely not worth the cost of another forward pass on the entire data, especially when the observation space and the model are complex.
Exploiting less stale data, using more epochs, severely degrades performance in both environments.

\begin{figure}[t]
\begin{center}
\centerline{\includegraphics[width=\columnwidth]{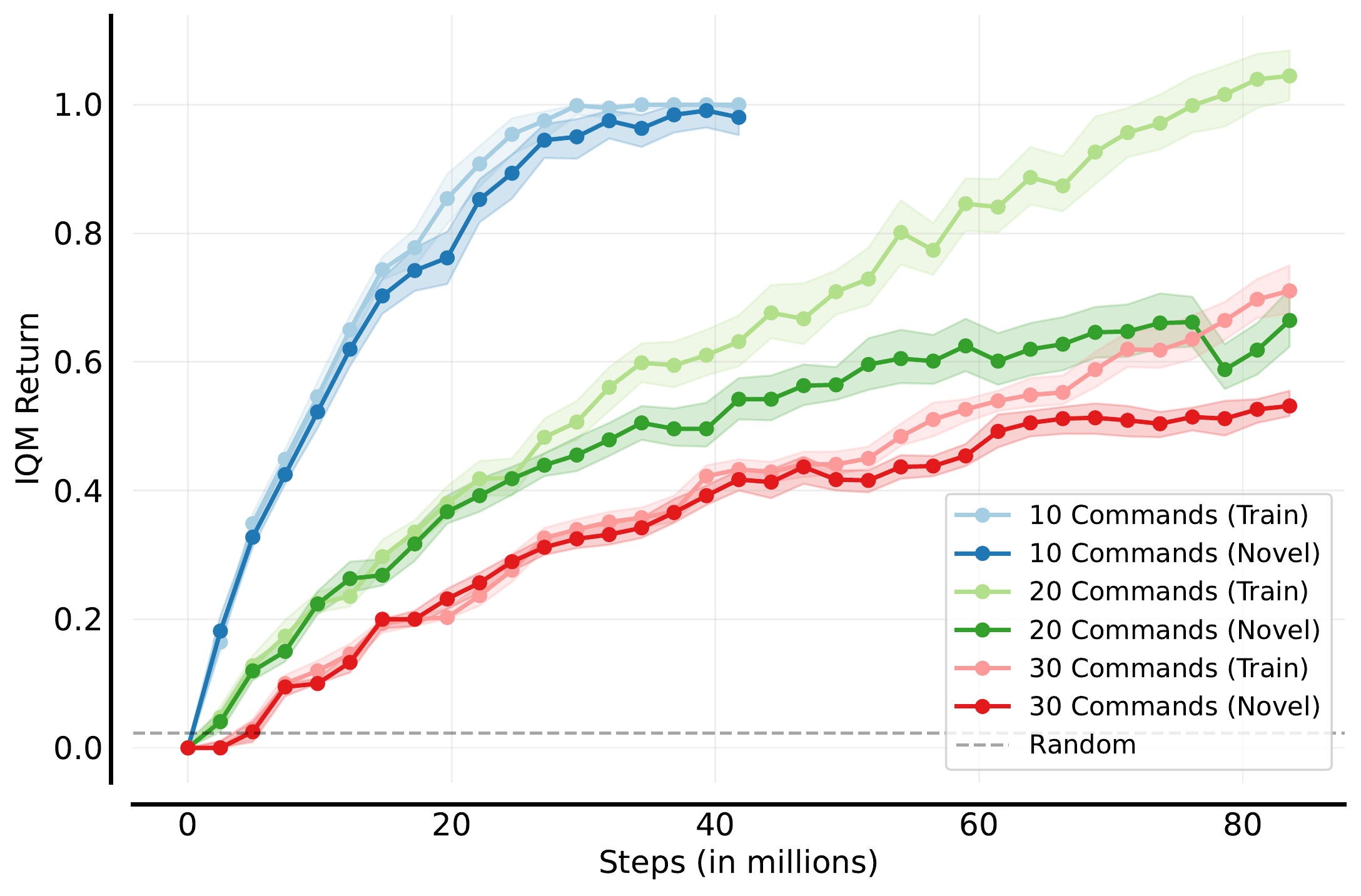}}
\caption{Generalization performance on training either 10, 20 or 30 commands in Mortar Mayhem.}
\label{fig:mortar_selected}
\end{center}
\vskip -0.2in
\end{figure}

\subsection{Observed Limits}

Reaching the limits of the underlying recurrent PPO implementation is not easily captured using Mortar Mayhem (Figure \ref{fig:mortar_selected}).
When trained on 20 or 30 commands, the highest possible return is $2.0$ or $3.0$.
Both variants do not fulfill this amplitude, but it can be considered that its solution is rather a question of model scale and computation time.
Concerning the ballet environment, its limitation to generalize is already apparent (Figure \ref{fig:ballet_selected}).
If trained on a constant number of dances and constant delay, the agent is able to generalize to novel seeds.
A completely different impression is obtained on Searing Spotlights.
The agent completely fails at the entire task, which is discussed in the next section.

\section{Discussion}

\subsection{Scaling the Difficulty of Mortar Mayhem}

As shown before, the agent needs memory to succeed in Mortar Mayhem.
The challenge of this environment is not just limited to the number of commands to be executed by the agent.
Applying modifications to Mortar Mayhem during training, like varying the number of commands and the delays between showing and executing each command, leads to a more complex distribution of training scenarios similar to the one seen in dancing the ballet.
Also, the reward signal can be made sparse, by only providing a reward upon successfully completing the entire command sequence.
Such alterations may further add value to challenging memory-based agents.

\subsection{Challenges of Searing Spotlights}

In Searing Spotlights, the agent is asked to collect all coins, while not being unveiled by any of the moving spotlights.
The agent's observation seldomly encompasses information on the whereabouts of the coins and itself, because these are only visible during the beginning of the episode and while being spotted by a light.
Therefore, this important information has to be inferred from past observations, which should be taken care of by the agent's memory.
Recurrent PPO does not seem to find any policy to solve this environment.
This can be due to the agent not being able to directly impact its observations for the most part of the episode, even though it perceives its last action.
So how should the agent be able to assign credit to its performed decisions?
The spotlights may worsen this dilemma by causing noise and distractions due to their stochastic behavior (random spawn positions and stochastic movement).
To gain more insights, the environment can be ablated using these simplifications:
\begin{compactitem}
    \item removing spotlight penalties (i.e. damage and reward)
    \item limiting the number of coins to one
    \item increasing the scale of the agent and the coin
    \item making the spotlights stationary
    \item making the agent or the coin permanently visible
\end{compactitem}
Applying the first two measures did not lead to successful training.
Scaling up the size of the agent and the coin by an order of two or three failed as well.
Once spotlights are stationary, the agent easily learns to collect the coin, but only when its scale is set to three\footnote{More figures, showing the scale of the agent and stationary spotlights, can be found in Appendix \ref{sec:simplifications}.}.
Lowering the scale to two does not always work out like before.
The original scale did not solve the task under stationary spotlights.
If the agent is scaled up and always visible, but the spotlights move, the task remains unsolved.
We did not explore all permutations of available simplifications yet, but it seems that moving spotlights are the toughest challenge in Searing Spotlights.
To overcome their threat of distraction, a completely different approach to recurrent PPO shall be considered.
We believe that a transformer architecture as episodic memory, like the Hierarchical Chunk Attention Memory \cite{Lampinen2021htm}, is more capable and robust to distractions.
Doing so, the agent could recall every single extracted feature from past observations to filter out spotlights and to better trace itself.
This, however, is left to future work.

\subsection{Staleness of Hidden States and Advantages}

Our observed results differ from the ones reported by 
\citeauthor{Kamarl2020} \yrcite{Kamarl2020} and \citeauthor{andrychowicz2021what} \yrcite{andrychowicz2021what}.
The monitored gradient norms and the Kullback–Leibler divergence between $\theta_{old}$ and $\theta$ do not yield further insights (Appendix \ref{sec:refresh_appendix}).
It seems that the issue of staleness in DRL needs further examination.
On-policy data might not suffer as much as off-policy ones from being stale.
This issue might be amplified in distributed training setups.
Maybe there is an impact on how advantage estimates are normalized.
\citeauthor{Chloe2020} \citeyear{Chloe2020} provide supplementary material where some agents performed worse on control tasks using normalization.
\citeauthor{Kamarl2020} \citeyear{Kamarl2020} normalized advantages across single episodes, while most implementations normalize those across minibatches or the entire batch of data.
It shall be investigated whether refreshing hidden states and advantages mitigate a higher variance when smaller chunks of data are used for normalization.

\subsection{Open Threads concerned with Recurrence}

More algorithmic details are left to future work.
One concerns the naive initialization of hidden states to zero.
Making the initial states learnable parameters as part of $\theta$ is non-trivial, because gradients are truncated and therefore cannot be backpropagated to the hidden states' origin.
\citeauthor{kapturowski2019r2d2} \citeyear{kapturowski2019r2d2} employed a 'burn-in' approach that produces a start hidden state by feeding the model a fragment of a sequence, while the remaining items of the sequence are used for optimization. 
However, it remains unclear how an agent shall initialize its hidden state when acting without the experience replay buffer used during training.
Moreover, future work can examine the utility of augmenting the training data by overlapping sequences, which was also touched by \citeauthor{kapturowski2019r2d2} \citeyear{kapturowski2019r2d2}.
To break impairing correlations \cite{Mnih2015}, which were reintroduced due to maintaining sequences, a stochastically positioned sliding window could be used to sample sequences.

\section{Conclusion}

In this paper, we demonstrated how to resolve the problems that originate from adding recurrence to the widely used PPO algorithm.
In more detail, these are concerned with the proper arrangement of the training data, the efficient organization of the neural net's forward pass, the correct sampling of hidden states related to sequence beginnings and the masking of paddings during loss computation.
Getting those right is essential to achieve an efficient and working implementation.

We further contributed the environments Mortar Mayhem and Searing Spotlights that challenge memory-based agents beyond solely capacity and distraction tasks.
These and the ballet environment are used to assess generalization and limitations of recurrent PPO.
Remarkably, when scaling up the number of training seeds in Mortar Mayhem, the agent is capable of generalizing well to novel seeds.
However, this is not observed in the ballet environment where the performance on training seeds is great, but worse than chance on novel ones.
Refreshing potentially stale hidden states and advantages did not improve results in our experiments.

At last, clear limitations of recurrent PPO are unveiled by trying to solve Searing Spotlights.
Only a heavily simplified version of this environment is solvable by the taken approach.
It seems that randomly moving spotlights cause too much noise and distraction to the agent.
To remedy the problems posed by Searing Spotlights, we see the potential for future work on adding an episodic memory to the agent via transformer architectures.

\bibliography{literature}
\bibliographystyle{icml2022}

\appendix
\newpage
\onecolumn

\icmltitle{Supplementary Material}

\section{Inhibiting Factors}
\label{sec:annex_bugs}

Deep Reinforcement Learning algorithms in general are prone to bugs, that are difficult to trace.
Adding recurrent layers to the PPO model notably increases the complexity of the entire implementation.
While working on the baseline implementation, several bugs severely hindered progress.
Two of those are related to the utilized machine learning framework PyTorch.
Unintentionally feeding $None$ as hidden state to the recurrent layer causes the network to initialize hidden states as zero.
This bug causes gradients to explode.
Further, when feeding a recurrent layer, PyTorch expects the sequence length to be the first dimension of the fed data.
Naively reshaping the data causes a catastrophic shift.
This bug may be circumvented by setting the batch first argument of the recurrent layer's forward function to true.
Also, one has to note that the output of an LSTM or GRU cell is already activated.
Adding another activation, like ReLU, on top might hurt performance.

\section{Mortar Mayhem}
\label{sec:annex_mortar}

\subsection{Configuration}

Mortar Mayhem can be configured differently using the following 5 options.
To further increase the challenge to the agent, one could vary those options during training as it has been done for the ballet environment.

\textbf{Number of commands to execute} (default: 10): The count of commands that the agent has to execute one by one after observing them.
\\
\textbf{Available commands} (default: first 5 choices): The command sequence can be composed of 9 different commands: Up, Down, Left, Right, Stay, Up Left, Up Right, Down Left and Down Right.
\\
\textbf{Command show duration} (default: 1 step): This determines for how long one command is presented to the agent.
\\
\textbf{Command show delay} (default: 1 step): The duration of not showing a command between commands.
\\
\textbf{Command execution delay} (default: 5 steps): This delay determines the number of steps the agent has to correctly execute the command.

\begin{figure}[b]
    \centering
    \subfigure[]{\label{fig:mm_unity_com}\includegraphics[width=\textwidth*\real{0.175}]{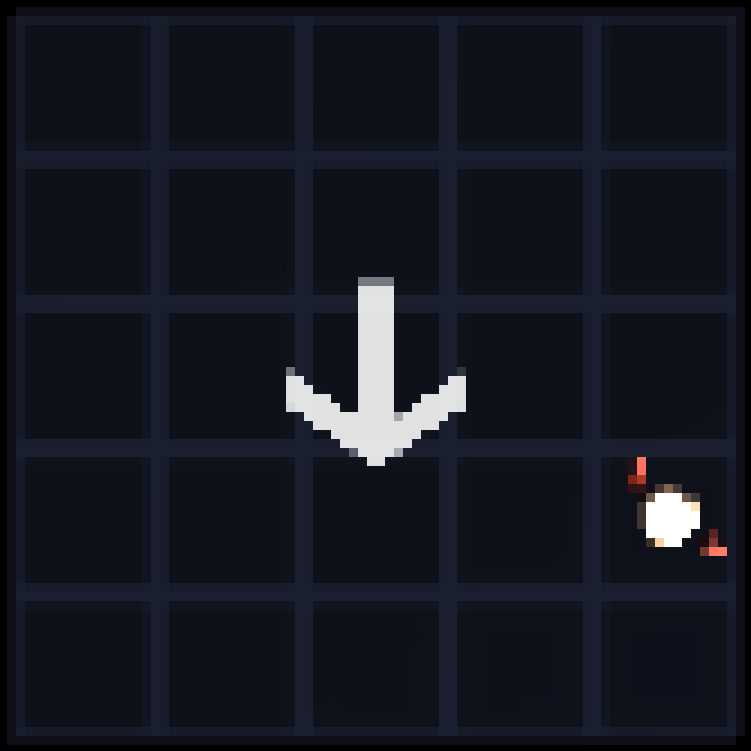}}
    \hspace{0.05\textwidth}
    \subfigure[]{\label{fig:mm_unity_alt}\includegraphics[width=\textwidth*\real{0.175}]{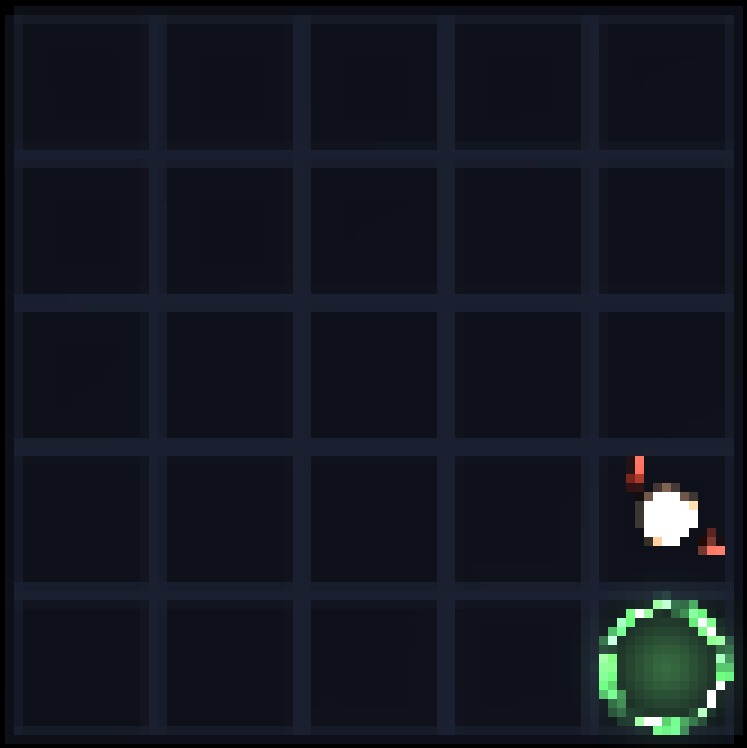}}
    \hspace{0.05\textwidth}
    \subfigure[]{\label{fig:mm_unity_exp}\includegraphics[width=\textwidth*\real{0.175}]{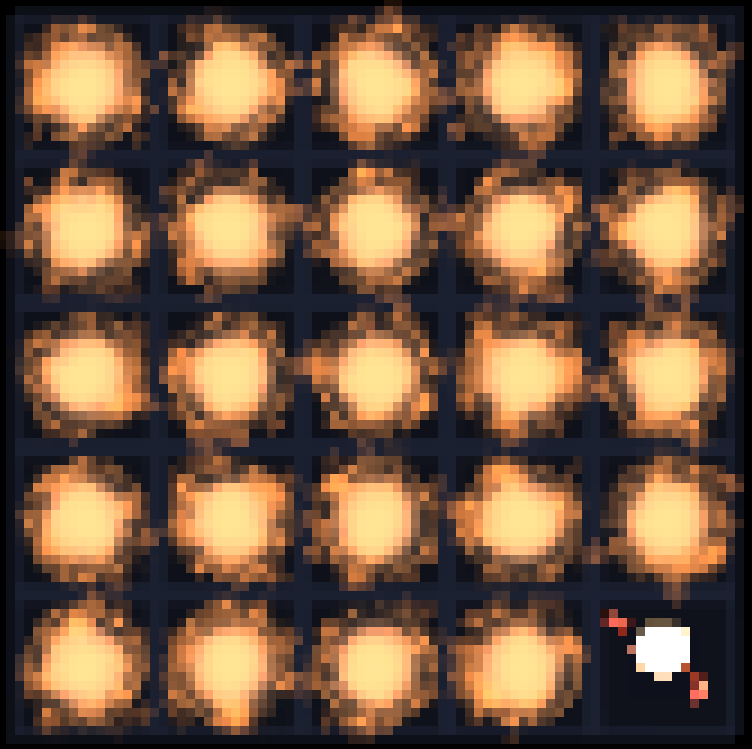}}
    \caption{The Unity ML-Agents version of Mortar Mayhem. a) shows how commands are presented to the agent, whereas b) presents the alternative visualization. The correct command execution is visualized using particle explosions (c)).}
    \label{fig:mm_unity}
\end{figure}

Also, an alternative version of the environment is implemented using the Unity ML-Agents toolkit \cite{Juliani2018mlagents} as shown by Figure \ref{fig:mm_unity}.
In this environment, the agent's actions lead to much smaller changes in position.
Hence, the agent needs more steps to position itself on an adjacent tile.
This environment features visual observations only.
Its action space is multi-discrete (3, 3) like how it is done in Searing Spotlights.
The first action dimension concerns the horizontal movement, whereas the other does the vertical movement.
Additionally, an alternative command visualization is available.
Instead of showing arrows, the target tile is highlighted using a particle effect.
The minimalistic gridworld version of this environment is used in this paper, because the Unity one cannot be easily scaled up to 32 environment instances.
Also, the simulation speed is faster.

\subsection{Example Episode}

\begin{figure}
    \centering
    \includegraphics[width=0.875\textwidth]{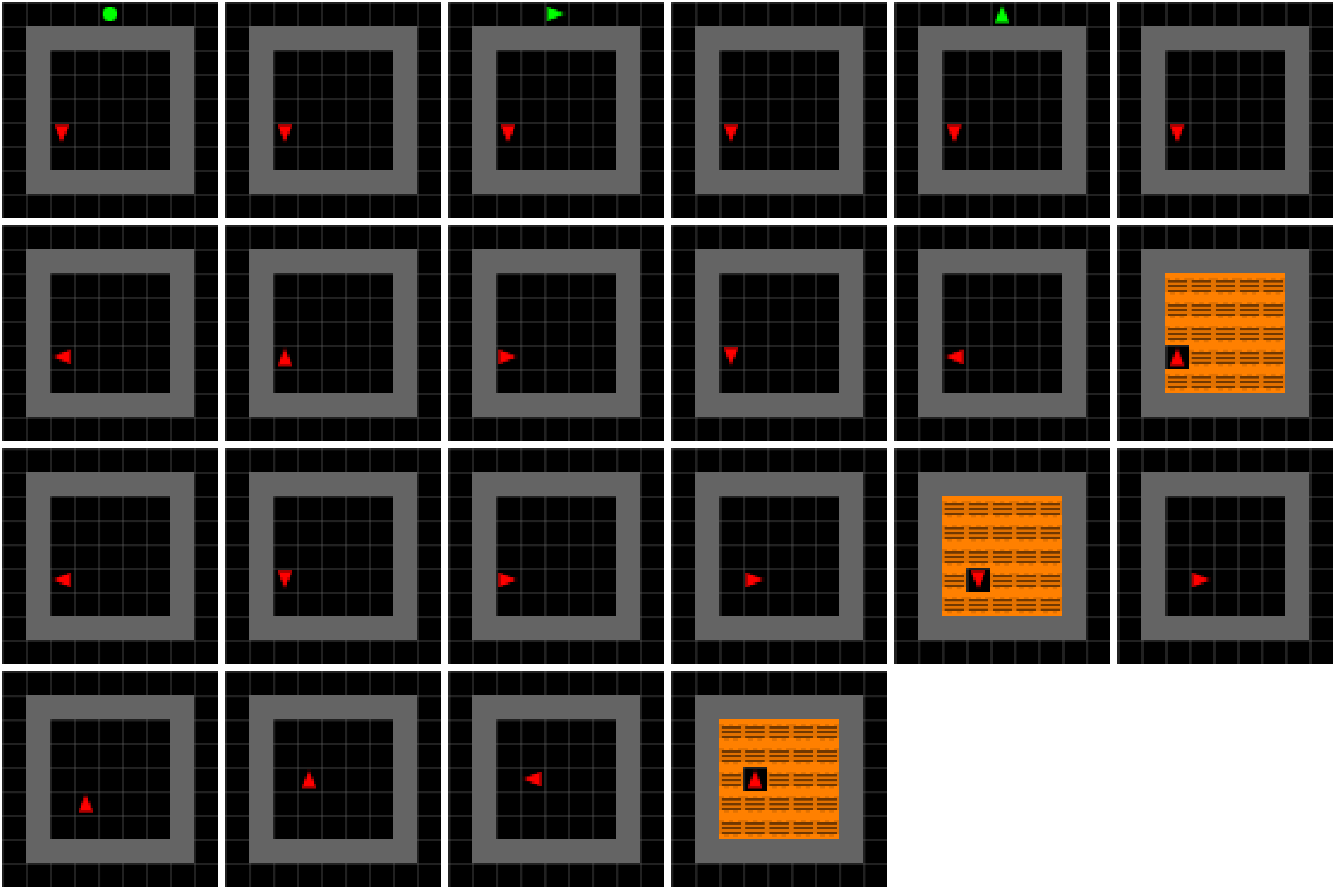}
    \caption{One complete episode of Mortar Mayhem using 3 commands.
    The first 6 frames demand the agent to memorize the commands stay, right and up.
    Once the 7th frame is reached, the agent's position and rotation is not frozen anymore.
    Its task is now to execute each command one by one.
    Success is signaled by a reward of $0.1$.
    Also, a visual feedback based on the lave tiles is provided.
    The final return of this episode is $0.3$.}
    \label{fig:mm_episode}
\end{figure}

Figure \ref{fig:mm_episode} shows the frames from left to right of an agent playing one episode of Mortar Mayhem given 3 commands.
During the first 6 frames, the agent has to observe the green commands, while not being able to move.
Then, the agent has 5 frames per command to correctly execute each one of them.
Given the command sequence from Figure \ref{fig:mm_episode}, the agent has to stay first, then to move right and finally to move up.

\section{Searing Spotlights}
\label{sec:annex_searing}

\subsection{Configuration}

Searing Spotlights can be configured differently using the following 18 options.

\textbf{Number of Spotlight Spawns} (default: 50). The number of spotlights that are spawned during one episode.
\\
\textbf{Initial Spawn Interval} (default: 12 steps). How many steps to wait for the next spotlight to be spawned. 
\\
\textbf{Interval Threshold} (default: 4 steps). The spawn interval decreases over time until it reaches its lover threshold.
\\
\textbf{Min Spotlight Speed} (default: 20). The minimum of the to be sampled spotlight speed.
\\
\textbf{Max Spotlight Speed} (default: 45). The maximum of the to be sampled spotlight speed.
\\
\textbf{Min Spotlight Radios} (default: 10). The minimum of the to be sampled spotlight radius.
\\
\textbf{Max Spotlight Radios} (default: 15). The maximum of the to be sampled spotlight radius.
\\
\textbf{Spotlight Damage} (default: 1). The amount of damage that one spotlight inflicts to the agent each step.
\\
\textbf{Light On Duration} (default: 3 steps). How many steps the global light is turned on during the beginning of the episode.
\\
\textbf{Light Dim Duration} (default: 10 steps). How many steps the global light is dimmed until off.
\\
\textbf{Number of Coins} (default: 1 coin). The number of coins that the agent has to collect.
\\
\textbf{Coin Scale} (default: 1). The scale of coins.
\\
\textbf{Coin Always Visible} (default: False). Whether the coin is never hidden in the dark.
\\
\textbf{Use Exit} (default: False). Whether the agent has to find an exit to positively terminate the environment. If coins are available, the agent has to collect those first.
\\
\textbf{Exit Scale} (default: 1). The scale of the exit.
\\
\textbf{Agent Speed} (default: 35). The speed of the agent to move around.
\\
\textbf{Agent Health} (default: 100). The amount of the agent's health, which is reduced by being visible inside spotlights. Once 0 is reached, the episode terminates.
\\
\textbf{Agent Always Visible} (default: False). Whether the agent is never hidden in the dark.

\subsection{Simplifications}
\label{sec:simplifications}

Like how it was mentioned in the main body of this paper, the following measures can be applied to simplify Searing Spotlights.

\begin{compactitem}
    \item removing spotlight penalties (i.e. damage and reward)
    \item limiting the number of coins to one
    \item increasing the scale of the agent and the coin
    \item making the spotlights stationary
    \item making the agent or the coin permanently visible
\end{compactitem}

\begin{figure}[H]
    \centering
    \subfigure[Scale 1]{\label{fig:ss_scale_1}\includegraphics[width=\textwidth*\real{0.175}]{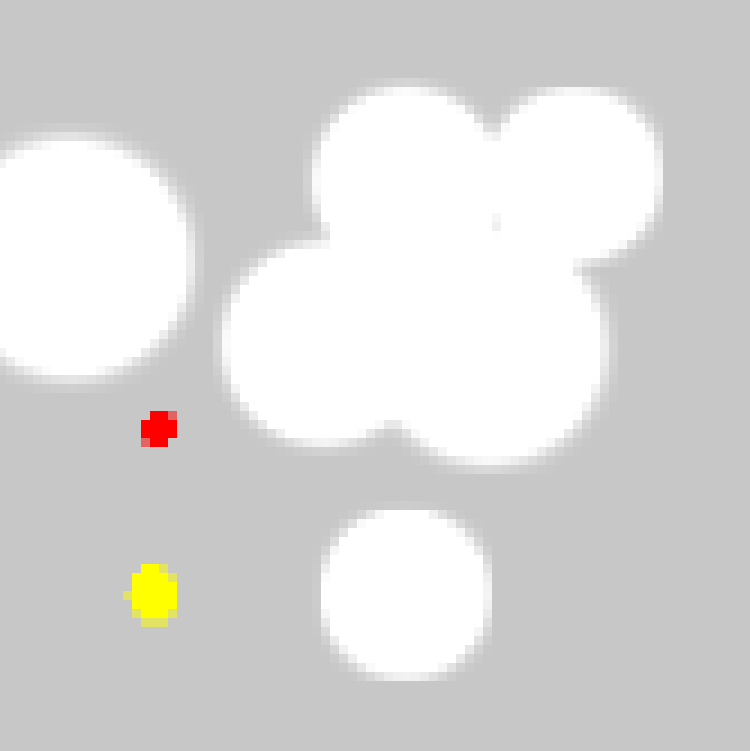}}
    \hspace{0.05\textwidth}
    \subfigure[Scale 2]{\label{fig:ss_scale_2}\includegraphics[width=\textwidth*\real{0.175}]{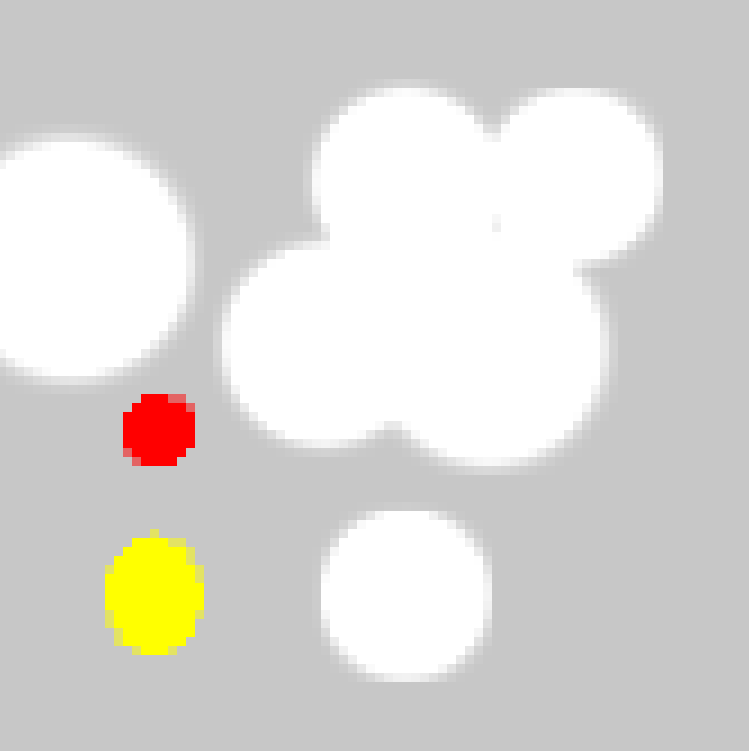}}
    \hspace{0.05\textwidth}
    \subfigure[Scale 3]{\label{fig:ss_scale_3}\includegraphics[width=\textwidth*\real{0.175}]{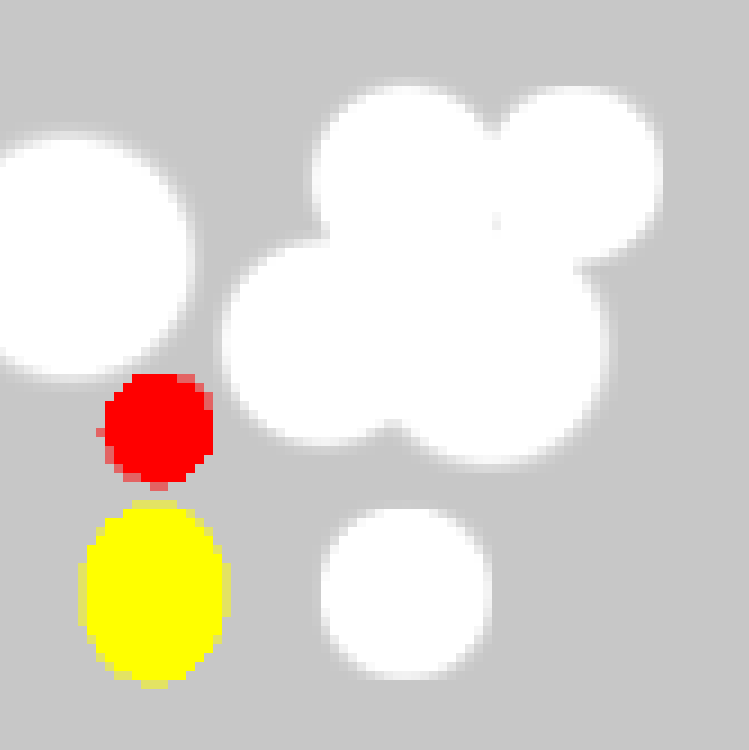}}
    \caption{The appearance of the agent (red dot) and the coin (yellow ellipsis) when scaled by 1, 2 or 3. These images (i.e. initial observations) feature 6 stationary spotlights.}
    \label{fig:ss_scale}
\end{figure}

Figure \ref{fig:ss_scale} describes the appearance of the environment upon scaling the coin and the agent.
When scaled up, the agent needs fewer steps to reach the coin.
Future work shall explore all possible permutations to analyze the exact level of solvability of Searing Spotlights.

\subsection{Example Episode}

\begin{figure}[hp!]
    \centering
    \includegraphics[width=0.875\textwidth]{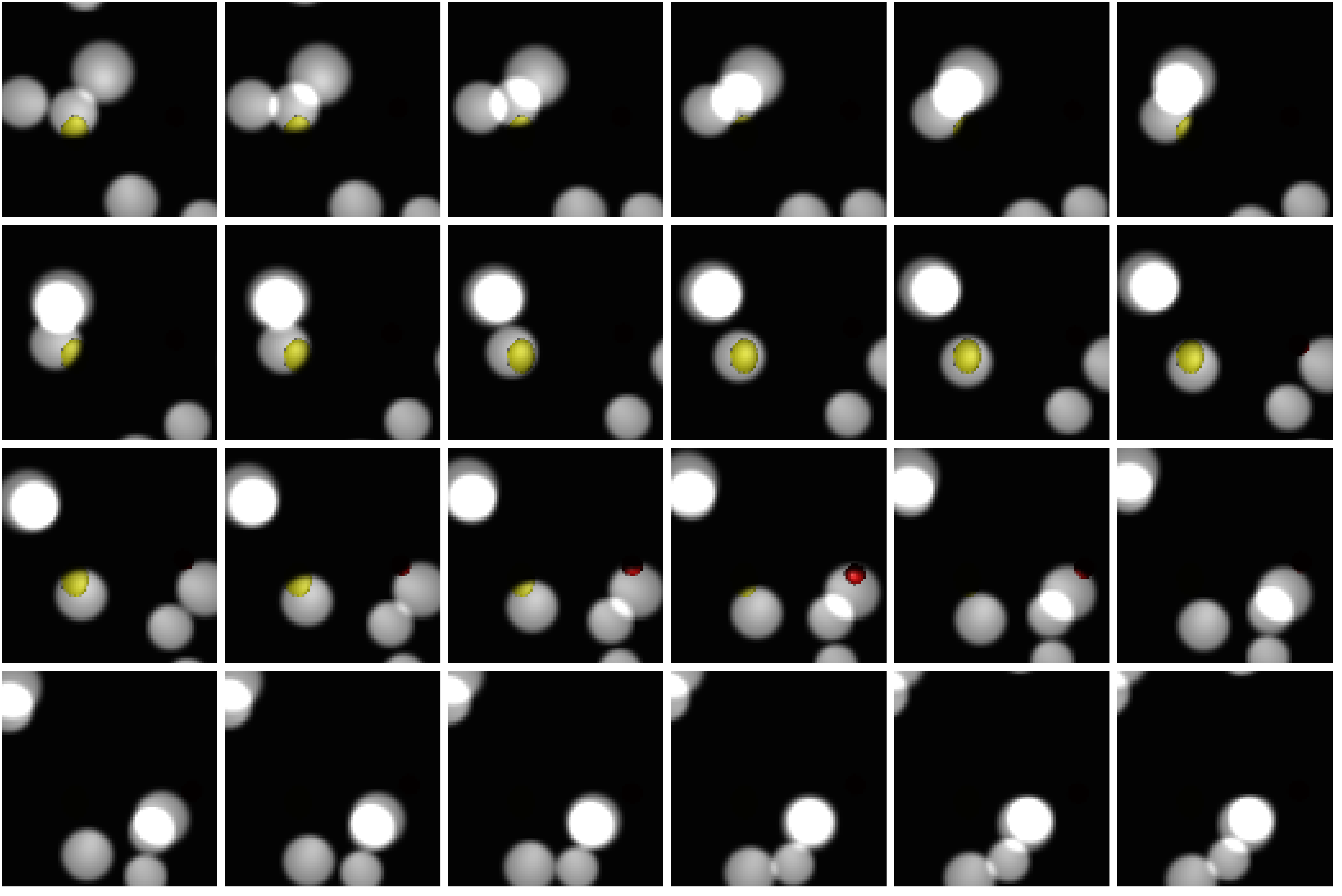}
    \caption{An episode segment of a random agent playing Searing Spotlights.
    About 6 spotlights stochastically move across the environment at this (advanced) point of time.
    Spotlights are spawned randomly on the outside of the environment.
    They move to a stochastically positioned and moving target on the opposite side.}
    \label{fig:ss_episode}
\end{figure}

One episode segment of Searing Spotlights is illustrated by Figure \ref{fig:ss_episode}.
The agent's observations are read from left to right.
The coin and the agent are at scale 2.

\subsection{Potential Strategies}

One strategy could be to ease memorizing the agent's position by moving continuously towards environment boundaries, because moving for long enough in one direction raises the confidence of being at the min or max position on one or both axes.
Another valid approach could be to trade-off between the spotlight's penalty and directly observing the agent's position.
The agent might intentionally risk being visible to reveal its position.

\clearpage
\section{Grid Search Results}
\label{sec:grid_search}

Grid search was conducted across various hyperparameters concerning recurrence on  Mortar Mayhem (Table \ref{tab:grid_mm}) and the ballet environment (Table \ref{tab:grid_ballet}).
The green shaded area corresponds to the recurrent layer type, the blue one to the sequence length and the red one to the hidden state size.
All possible permutations were trained three times and evaluated on 5 training and 5 novel seeds.
Each data point accumulates the episodes of the last three checkpoints.
The results show that longer sequences and greater hidden state sizes are favorable.
GRU is usually superior to LSTM.

\begin{table*}[h]
\centering
\caption{Mortar Mayhem Grid Search Results. Each column shows the percentage of succeeding at least on the specified number of commands. Note: 200 seeds were only present during training, which limit the agent's ability to generalize. The training lasted for about 27 million steps.}
\vskip 0.1in
\begin{small}
\begin{tabular}{|l|rrrrrrrrr|rrrrrrr|}
\hline
        & \multicolumn{9}{c|}{Training Seeds}                                                                                                                                                                                                                                                                                                                     & \multicolumn{7}{c|}{Novel Seeds}                                                                                                                                                                                                                                                                                              \\ \hline
\# Com. & \multicolumn{1}{l}{\cellcolor[HTML]{EFEFEF}1} & \multicolumn{1}{l}{2} & \multicolumn{1}{l}{\cellcolor[HTML]{EFEFEF}3} & \multicolumn{1}{l}{4} & \multicolumn{1}{l}{\cellcolor[HTML]{EFEFEF}5} & \multicolumn{1}{l}{6} & \multicolumn{1}{l}{\cellcolor[HTML]{EFEFEF}7} & \multicolumn{1}{l}{8} & \multicolumn{1}{l|}{\cellcolor[HTML]{EFEFEF}9} & \multicolumn{1}{l}{1} & \multicolumn{1}{l}{\cellcolor[HTML]{EFEFEF}2} & \multicolumn{1}{l}{3} & \multicolumn{1}{l}{\cellcolor[HTML]{EFEFEF}4} & \multicolumn{1}{l}{5} & \multicolumn{1}{l}{\cellcolor[HTML]{EFEFEF}6} & \multicolumn{1}{l|}{7}  \\ \hline
\rowcolor[HTML]{CCECE3} 
GRU     & \cellcolor[HTML]{C5E5DC}0.92                  & 0.82                  & \cellcolor[HTML]{C5E5DC}0.71                  & 0.64                  & \cellcolor[HTML]{C5E5DC}0.55                  & 0.5                   & \cellcolor[HTML]{C5E5DC}0.4                   & 0.29                  & \cellcolor[HTML]{C5E5DC}0.24                   & 0.51                  & \cellcolor[HTML]{C5E5DC}0.22                  & 0.11                  & \cellcolor[HTML]{C5E5DC}0.06                  & 0.003                 & \cellcolor[HTML]{C5E5DC}0.0                   & 0.0              \\
\rowcolor[HTML]{CCECE3} 
LSTM    & \cellcolor[HTML]{C5E5DC}0.92                  & 0.8                   & \cellcolor[HTML]{C5E5DC}0.66                  & 0.61                  & \cellcolor[HTML]{C5E5DC}0.51                  & 0.45                  & \cellcolor[HTML]{C5E5DC}0.38                  & 0.27                  & \cellcolor[HTML]{C5E5DC}0.21                   & 0.42                  & \cellcolor[HTML]{C5E5DC}0.1                   & 0.05                  & \cellcolor[HTML]{C5E5DC}0.03                  & 0.001                 & \cellcolor[HTML]{C5E5DC}0.0                   & 0.0               \\ \hline
\rowcolor[HTML]{CCE3F0} 
1       & \cellcolor[HTML]{C3DAE5}0.77                  & 0.46                  & \cellcolor[HTML]{C3DAE5}0.29                  & 0.23                  & \cellcolor[HTML]{C3DAE5}0.17                  & 0.1                   & \cellcolor[HTML]{C3DAE5}0.03                  & 0.0                   & \cellcolor[HTML]{C3DAE5}0.0                    & 0.23                  & \cellcolor[HTML]{C3DAE5}0.04                  & 0.01                  & \cellcolor[HTML]{C3DAE5}0.01                  & 0.0                   & \cellcolor[HTML]{C3DAE5}0.0                   & 0.0               \\
\rowcolor[HTML]{CCE3F0} 
8       & \cellcolor[HTML]{C3DAE5}0.93                  & 0.82                  & \cellcolor[HTML]{C3DAE5}0.61                  & 0.53                  & \cellcolor[HTML]{C3DAE5}0.36                  & 0.26                  & \cellcolor[HTML]{C3DAE5}0.14                  & 0.03                  & \cellcolor[HTML]{C3DAE5}0.02                   & 0.28                  & \cellcolor[HTML]{C3DAE5}0.06                  & 0.01                  & \cellcolor[HTML]{C3DAE5}0.003                 & 0.0                   & \cellcolor[HTML]{C3DAE5}0.0                   & 0.0                \\
\rowcolor[HTML]{CCE3F0} 
16      & \cellcolor[HTML]{C3DAE5}0.87                  & 0.78                  & \cellcolor[HTML]{C3DAE5}0.61                  & 0.55                  & \cellcolor[HTML]{C3DAE5}0.45                  & 0.39                  & \cellcolor[HTML]{C3DAE5}0.27                  & 0.14                  & \cellcolor[HTML]{C3DAE5}0.11                   & 0.3                   & \cellcolor[HTML]{C3DAE5}0.06                  & 0.03                  & \cellcolor[HTML]{C3DAE5}0.02                  & 0.001                 & \cellcolor[HTML]{C3DAE5}0.0                   & 0.0                  \\
\rowcolor[HTML]{CCE3F0} 
32      & \cellcolor[HTML]{C3DAE5}0.99                  & 0.92                  & \cellcolor[HTML]{C3DAE5}0.84                  & 0.78                  & \cellcolor[HTML]{C3DAE5}0.67                  & 0.65                  & \cellcolor[HTML]{C3DAE5}0.57                  & 0.4                   & \cellcolor[HTML]{C3DAE5}0.29                   & 0.69                  & \cellcolor[HTML]{C3DAE5}0.3                   & 0.15                  & \cellcolor[HTML]{C3DAE5}0.08                  & 0.001                 & \cellcolor[HTML]{C3DAE5}0.0                   & 0.0                 \\
\rowcolor[HTML]{CCE3F0} 
64      & \cellcolor[HTML]{C3DAE5}0.99                  & 0.95                  & \cellcolor[HTML]{C3DAE5}0.89                  & 0.84                  & \cellcolor[HTML]{C3DAE5}0.77                  & 0.74                  & \cellcolor[HTML]{C3DAE5}0.68                  & 0.54                  & \cellcolor[HTML]{C3DAE5}0.45                   & 0.63                  & \cellcolor[HTML]{C3DAE5}0.23                  & 0.13                  & \cellcolor[HTML]{C3DAE5}0.09                  & 0.004                 & \cellcolor[HTML]{C3DAE5}0.0                   & 0.0                  \\
\rowcolor[HTML]{CCE3F0} 
Max     & \cellcolor[HTML]{C3DAE5}0.99                  & 0.92                  & \cellcolor[HTML]{C3DAE5}0.87                  & 0.82                  & \cellcolor[HTML]{C3DAE5}0.76                  & 0.72                  & \cellcolor[HTML]{C3DAE5}0.67                  & 0.55                  & \cellcolor[HTML]{C3DAE5}0.49                   & 0.65                  & \cellcolor[HTML]{C3DAE5}0.27                  & 0.16                  & \cellcolor[HTML]{C3DAE5}0.09                  & 0.006                 & \cellcolor[HTML]{C3DAE5}0.0                   & 0.0                  \\ \hline
\rowcolor[HTML]{F5E4F2} 
64      & \cellcolor[HTML]{EDDDEA}0.9                   & 0.7                   & \cellcolor[HTML]{EDDDEA}0.5                   & 0.41                  & \cellcolor[HTML]{EDDDEA}0.27                  & 0.22                  & \cellcolor[HTML]{EDDDEA}0.1                   & 0.03                  & \cellcolor[HTML]{EDDDEA}0.02                   & 0.44                  & \cellcolor[HTML]{EDDDEA}0.14                  & 0.08                  & \cellcolor[HTML]{EDDDEA}0.04                  & 0.001                 & \cellcolor[HTML]{EDDDEA}0.0                   & 0.0                \\
\rowcolor[HTML]{F5E4F2} 
128     & \cellcolor[HTML]{EDDDEA}0.9                   & 0.77                  & \cellcolor[HTML]{EDDDEA}0.63                  & 0.57                  & \cellcolor[HTML]{EDDDEA}0.45                  & 0.38                  & \cellcolor[HTML]{EDDDEA}0.25                  & 0.14                  & \cellcolor[HTML]{EDDDEA}0.09                   & 0.41                  & \cellcolor[HTML]{EDDDEA}0.1                   & 0.06                  & \cellcolor[HTML]{EDDDEA}0.04                  & 0.002                 & \cellcolor[HTML]{EDDDEA}0.0                   & 0.0               \\
\rowcolor[HTML]{F5E4F2} 
256     & \cellcolor[HTML]{EDDDEA}0.92                  & 0.81                  & \cellcolor[HTML]{EDDDEA}0.7                   & 0.63                  & \cellcolor[HTML]{EDDDEA}0.54                  & 0.5                   & \cellcolor[HTML]{EDDDEA}0.44                  & 0.3                   & \cellcolor[HTML]{EDDDEA}0.23                   & 0.47                  & \cellcolor[HTML]{EDDDEA}0.14                  & 0.07                  & \cellcolor[HTML]{EDDDEA}0.04                  & 0.002                 & \cellcolor[HTML]{EDDDEA}0.0                   & 0.0               \\
\rowcolor[HTML]{F5E4F2} 
512     & \cellcolor[HTML]{EDDDEA}0.94                  & 0.88                  & \cellcolor[HTML]{EDDDEA}0.79                  & 0.74                  & \cellcolor[HTML]{EDDDEA}0.68                  & 0.63                  & \cellcolor[HTML]{EDDDEA}0.57                  & 0.46                  & \cellcolor[HTML]{EDDDEA}0.39                   & 0.48                  & \cellcolor[HTML]{EDDDEA}0.18                  & 0.08                  & \cellcolor[HTML]{EDDDEA}0.05                  & 0.003                 & \cellcolor[HTML]{EDDDEA}0.0                   & 0.0              \\
\rowcolor[HTML]{F5E4F2} 
1024    & \cellcolor[HTML]{EDDDEA}0.95                  & 0.88                  & \cellcolor[HTML]{EDDDEA}0.81                  & 0.77                  & \cellcolor[HTML]{EDDDEA}0.71                  & 0.65                  & \cellcolor[HTML]{EDDDEA}0.59                  & 0.46                  & \cellcolor[HTML]{EDDDEA}0.39                   & 0.52                  & \cellcolor[HTML]{EDDDEA}0.22                  & 0.12                  & \cellcolor[HTML]{EDDDEA}0.07                  & 0.002                 & \cellcolor[HTML]{EDDDEA}0.0                   & 0.0               \\ \hline
\end{tabular}
\end{small}
\label{tab:grid_mm}
\end{table*}

\begin{table}[h!]
\centering
\caption{Dancing the Ballet Grid Search Results. Each column shows the percentage of succeeding at 2, 4 or 8 dances. Note: 500 seeds were present during training. The training lasted for about 27 million steps.}
\vskip 0.1in
\begin{small}
\begin{tabular}{|l|rrr|rrr|}
\hline
        & \multicolumn{3}{c|}{Training Seeds}                                                                                             & \multicolumn{3}{c|}{Novel Seeds}                                                                      \\ \hline
\# Dances & \multicolumn{1}{l}{\cellcolor[HTML]{EFEFEF}2} & \multicolumn{1}{l}{4} & \multicolumn{1}{l|}{\cellcolor[HTML]{EFEFEF}8} & \multicolumn{1}{l}{2} & \multicolumn{1}{l}{\cellcolor[HTML]{EFEFEF}4} & \multicolumn{1}{l|}{8} \\ \hline
\rowcolor[HTML]{CCECE3} 
GRU     & \cellcolor[HTML]{C5E5DC}0.96                  & 0.91                  & \cellcolor[HTML]{C5E5DC}0.81                   & 0.6                   & \cellcolor[HTML]{C5E5DC}0.23                  & 0.17                   \\
\rowcolor[HTML]{CCECE3} 
LSTM    & \cellcolor[HTML]{C5E5DC}0.96                  & 0.81                  & \cellcolor[HTML]{C5E5DC}0.66                   & 0.59                  & \cellcolor[HTML]{C5E5DC}0.26                  & 0.15                   \\ \hline
\rowcolor[HTML]{CCE3F0} 
1       & \cellcolor[HTML]{C3DAE5}0.93                  & 0.8                   & \cellcolor[HTML]{C3DAE5}0.64                   & 0.6                   & \cellcolor[HTML]{C3DAE5}0.24                  & 0.16                   \\
\rowcolor[HTML]{CCE3F0} 
8       & \cellcolor[HTML]{C3DAE5}0.97                  & 0.82                  & \cellcolor[HTML]{C3DAE5}0.72                   & 0.59                  & \cellcolor[HTML]{C3DAE5}0.24                  & 0.17                   \\
\rowcolor[HTML]{CCE3F0} 
16      & \cellcolor[HTML]{C3DAE5}0.95                  & 0.88                  & \cellcolor[HTML]{C3DAE5}0.72                   & 0.57                  & \cellcolor[HTML]{C3DAE5}0.24                  & 0.14                   \\
\rowcolor[HTML]{CCE3F0} 
32      & \cellcolor[HTML]{C3DAE5}0.96                  & 0.87                  & \cellcolor[HTML]{C3DAE5}0.77                   & 0.6                   & \cellcolor[HTML]{C3DAE5}0.24                  & 0.17                   \\
\rowcolor[HTML]{CCE3F0} 
64      & \cellcolor[HTML]{C3DAE5}0.97                  & 0.9                   & \cellcolor[HTML]{C3DAE5}0.77                   & 0.62                  & \cellcolor[HTML]{C3DAE5}0.28                  & 0.17                   \\
\rowcolor[HTML]{CCE3F0} 
Max       & \cellcolor[HTML]{C3DAE5}0.97                  & 0.89                  & \cellcolor[HTML]{C3DAE5}0.8                    & 0.59                  & \cellcolor[HTML]{C3DAE5}0.27                  & 0.17                   \\ \hline
\rowcolor[HTML]{F5E4F2} 
64      & \cellcolor[HTML]{EDDDEA}0.95                  & 0.82                  & \cellcolor[HTML]{EDDDEA}0.68                   & 0.6                   & \cellcolor[HTML]{EDDDEA}0.2                   & 0.16                   \\
\rowcolor[HTML]{F5E4F2} 
128     & \cellcolor[HTML]{EDDDEA}0.96                  & 0.85                  & \cellcolor[HTML]{EDDDEA}0.76                   & 0.62                  & \cellcolor[HTML]{EDDDEA}0.24                  & 0.16                   \\
\rowcolor[HTML]{F5E4F2} 
256     & \cellcolor[HTML]{EDDDEA}0.97                  & 0.9                   & \cellcolor[HTML]{EDDDEA}0.74                   & 0.58                  & \cellcolor[HTML]{EDDDEA}0.25                  & 0.18                   \\
\rowcolor[HTML]{F5E4F2} 
512     & \cellcolor[HTML]{EDDDEA}0.96                  & 0.85                  & \cellcolor[HTML]{EDDDEA}0.72                   & 0.59                  & \cellcolor[HTML]{EDDDEA}0.27                  & 0.16                   \\
\rowcolor[HTML]{F5E4F2} 
1024    & \cellcolor[HTML]{EDDDEA}0.95                  & 0.89                  & \cellcolor[HTML]{EDDDEA}0.79                   & 0.59                  & \cellcolor[HTML]{EDDDEA}0.28                  & 0.16                   \\ \hline
\end{tabular}
\end{small}
\label{tab:grid_ballet}
\end{table}

\clearpage
\section{Refreshing Stale Hidden States and Advantages}
\label{sec:refresh_appendix}

\begin{figure}[hp!]
    \centering
    \subfigure[]{\label{fig:mm_model_norm}\includegraphics[width=\textwidth*\real{0.33}]{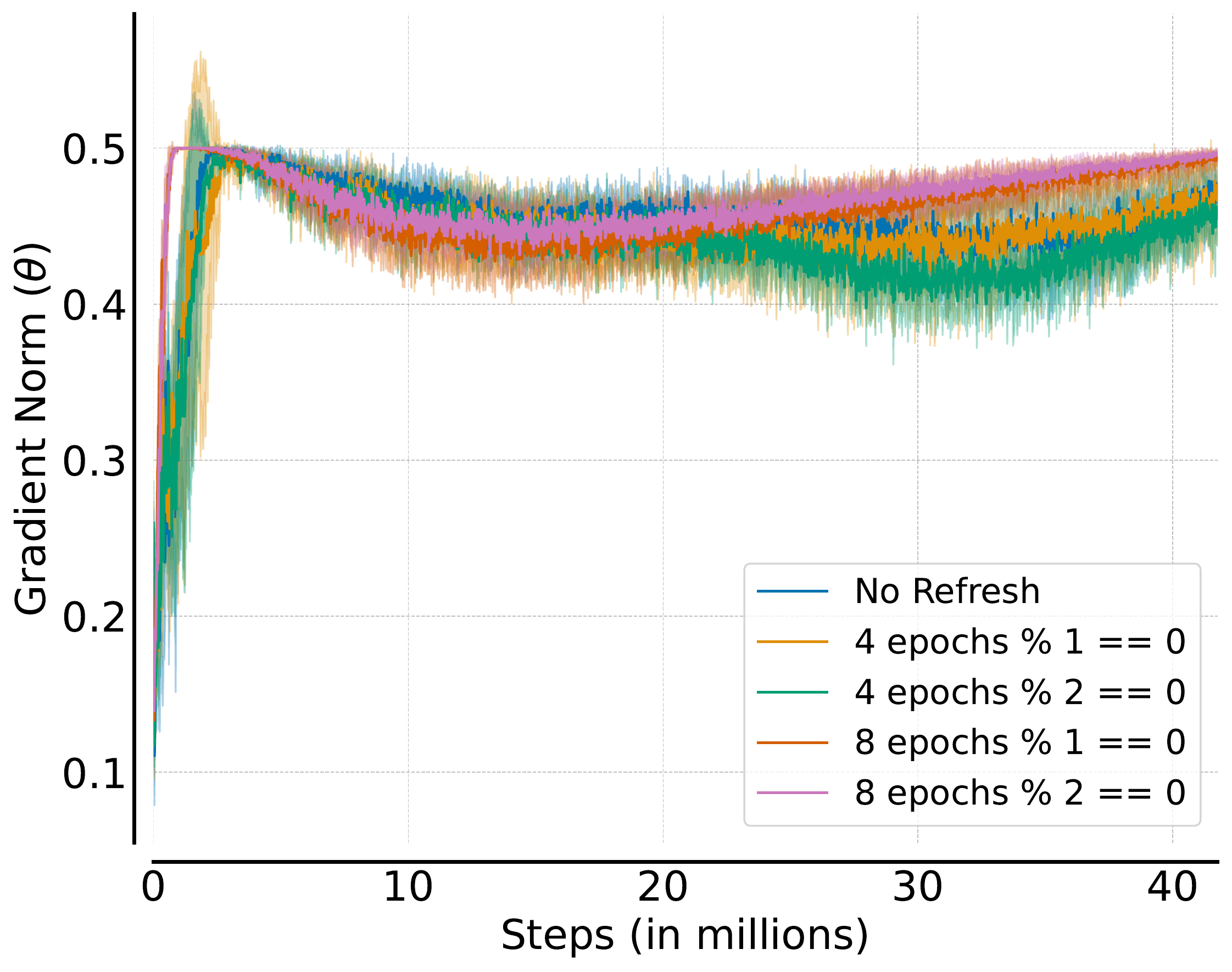}}
    \subfigure[]{\label{fig:mm_rec_norm}\includegraphics[width=\textwidth*\real{0.33}]{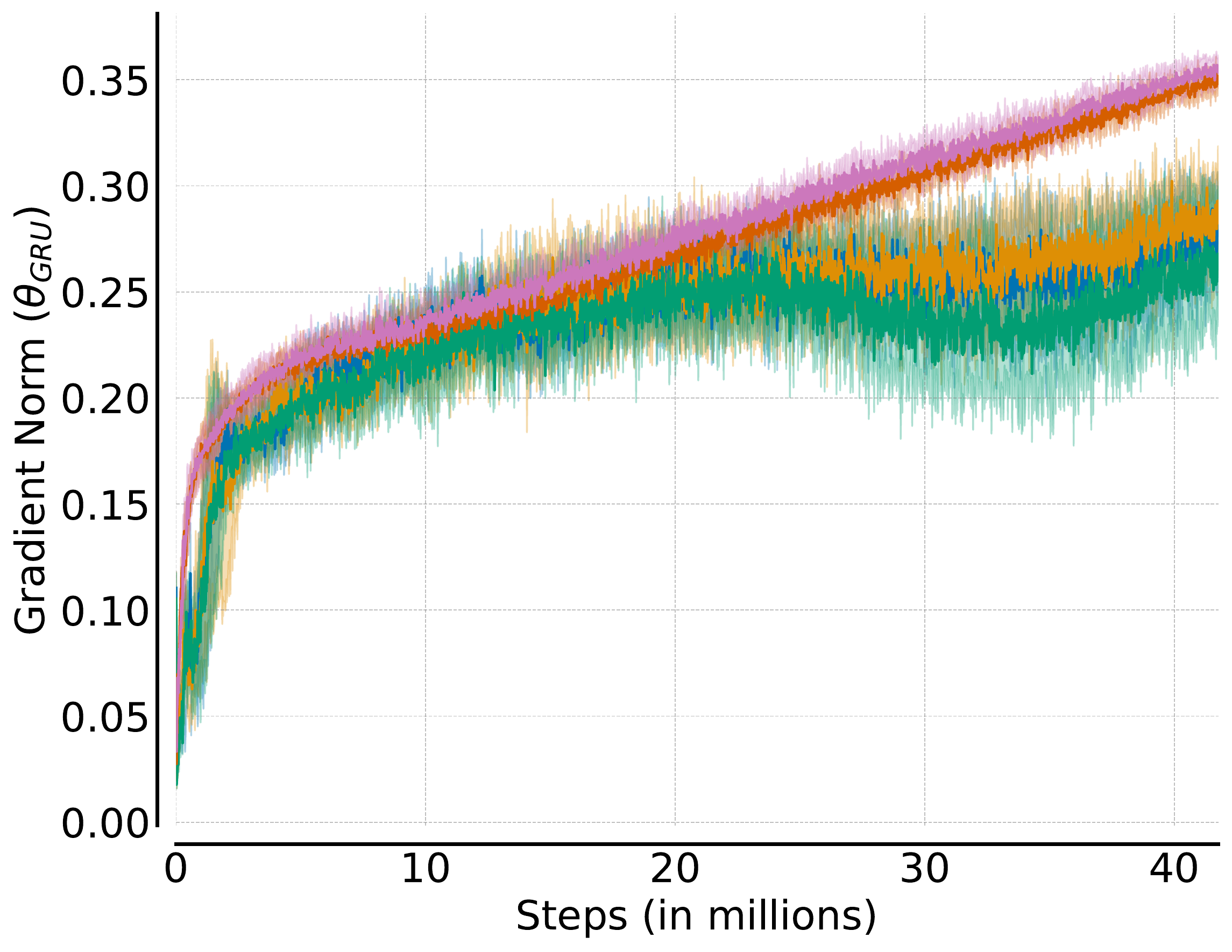}}
    \subfigure[]{\label{fig:mm_kl_divergence}\includegraphics[width=\textwidth*\real{0.33}]{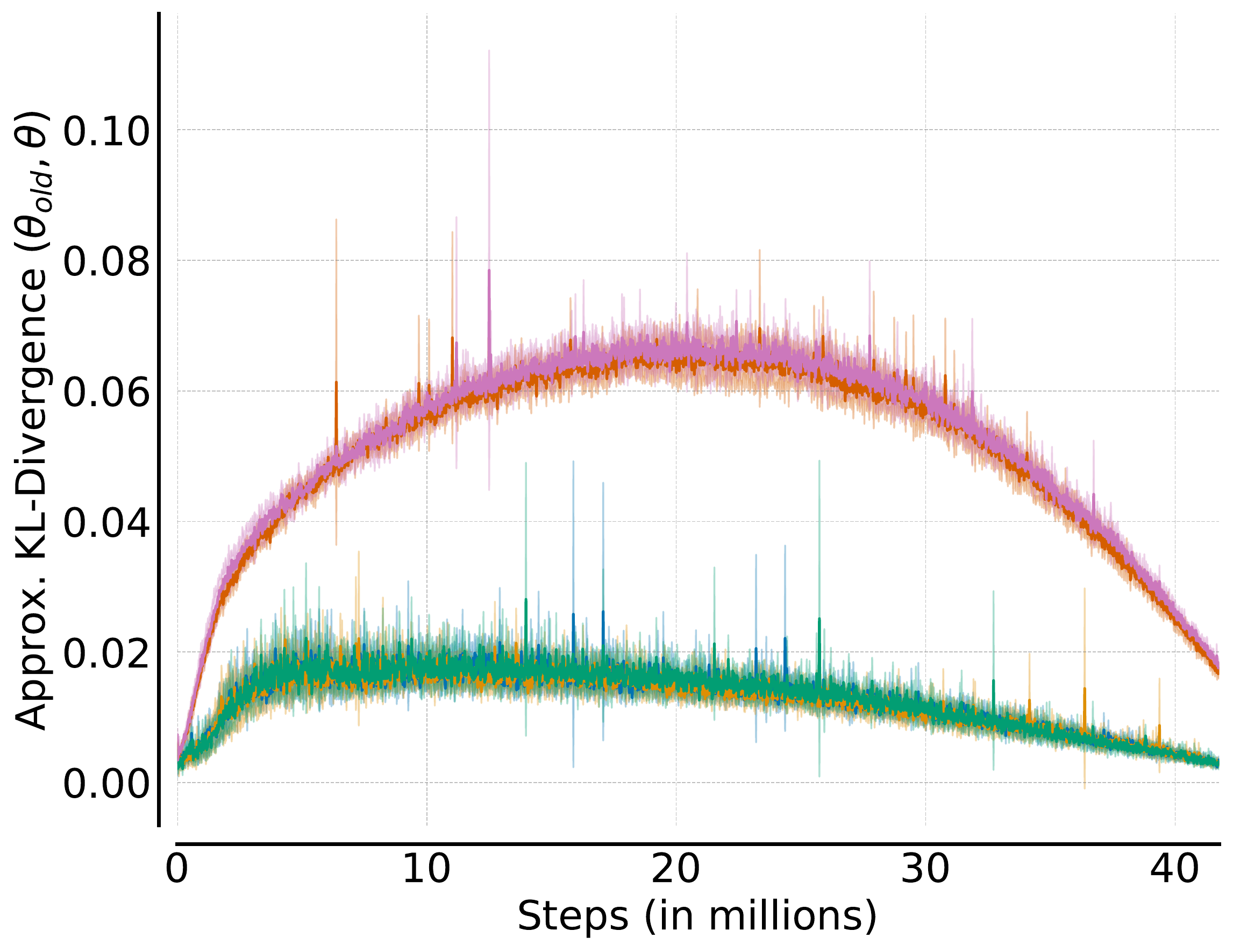}}
    \caption{Monitoring of the training runs in Mortar Mayhem when refreshing stale hidden states and advantages. a) shows the gradient norm of all model parameters, whereas b) presents the gradient norm only for parameters of the recurrent layer GRU. The Kullback–Leibler divergence between $\theta_{old}$ and $\theta$ is illustrated by c). All curves show the mean and the standard deviation.}
    \label{fig:mm_refresh_analysis}
\end{figure}

\begin{figure}[hp!]
    \centering
    \subfigure[]{\label{fig:ballet_model_norm}\includegraphics[width=\textwidth*\real{0.33}]{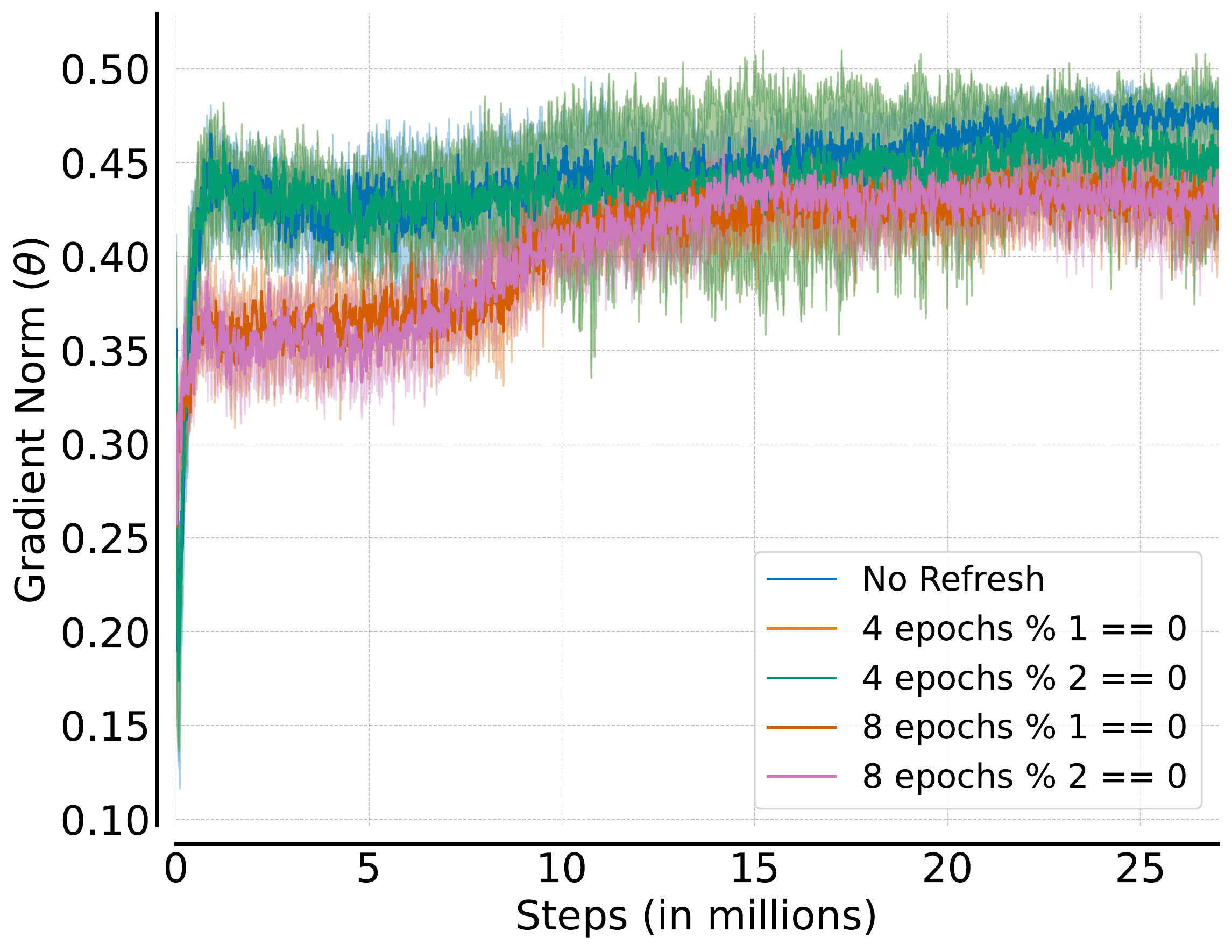}}
    \subfigure[]{\label{fig:ballet_rec_norm}\includegraphics[width=\textwidth*\real{0.33}]{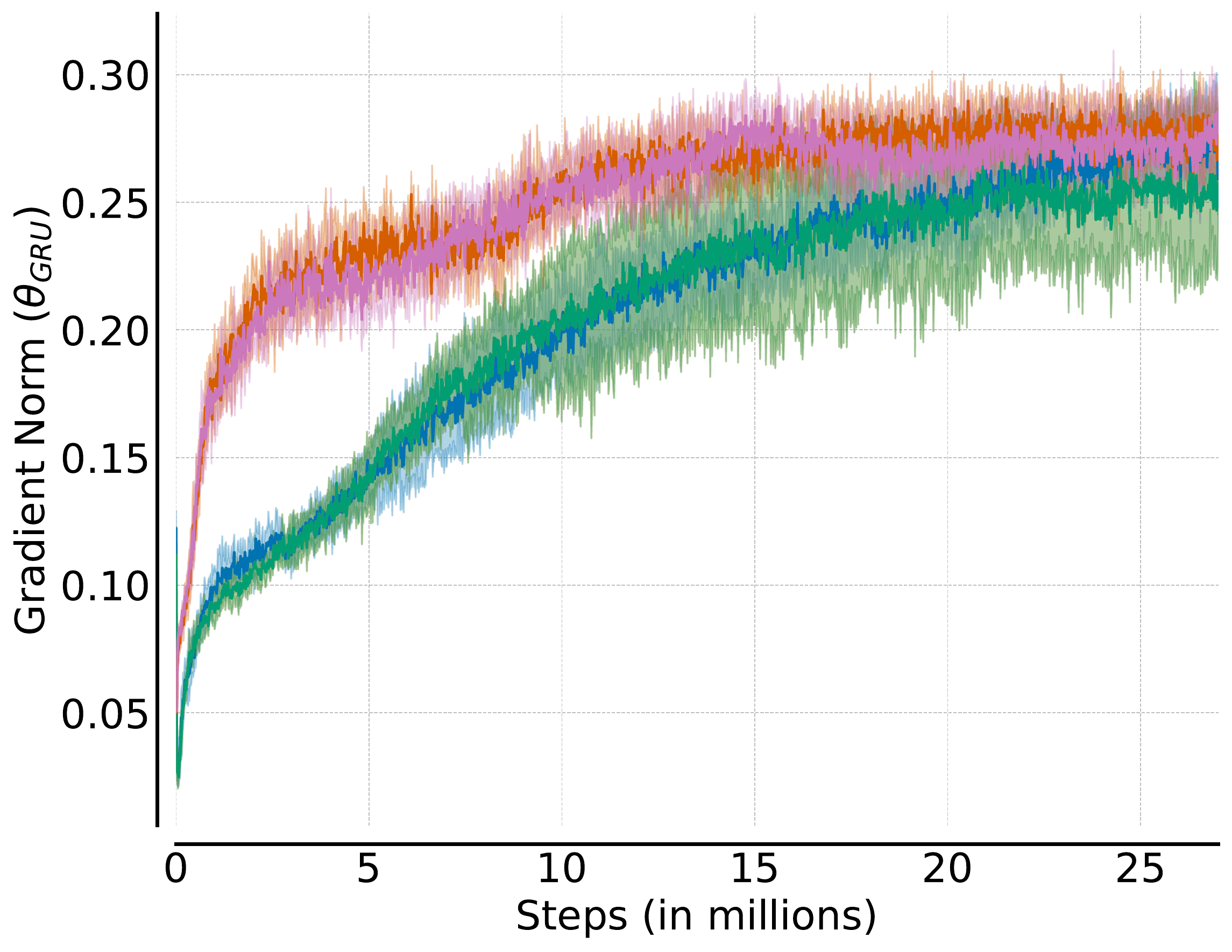}}
    \subfigure[]{\label{fig:ballet_kl_divergence}\includegraphics[width=\textwidth*\real{0.33}]{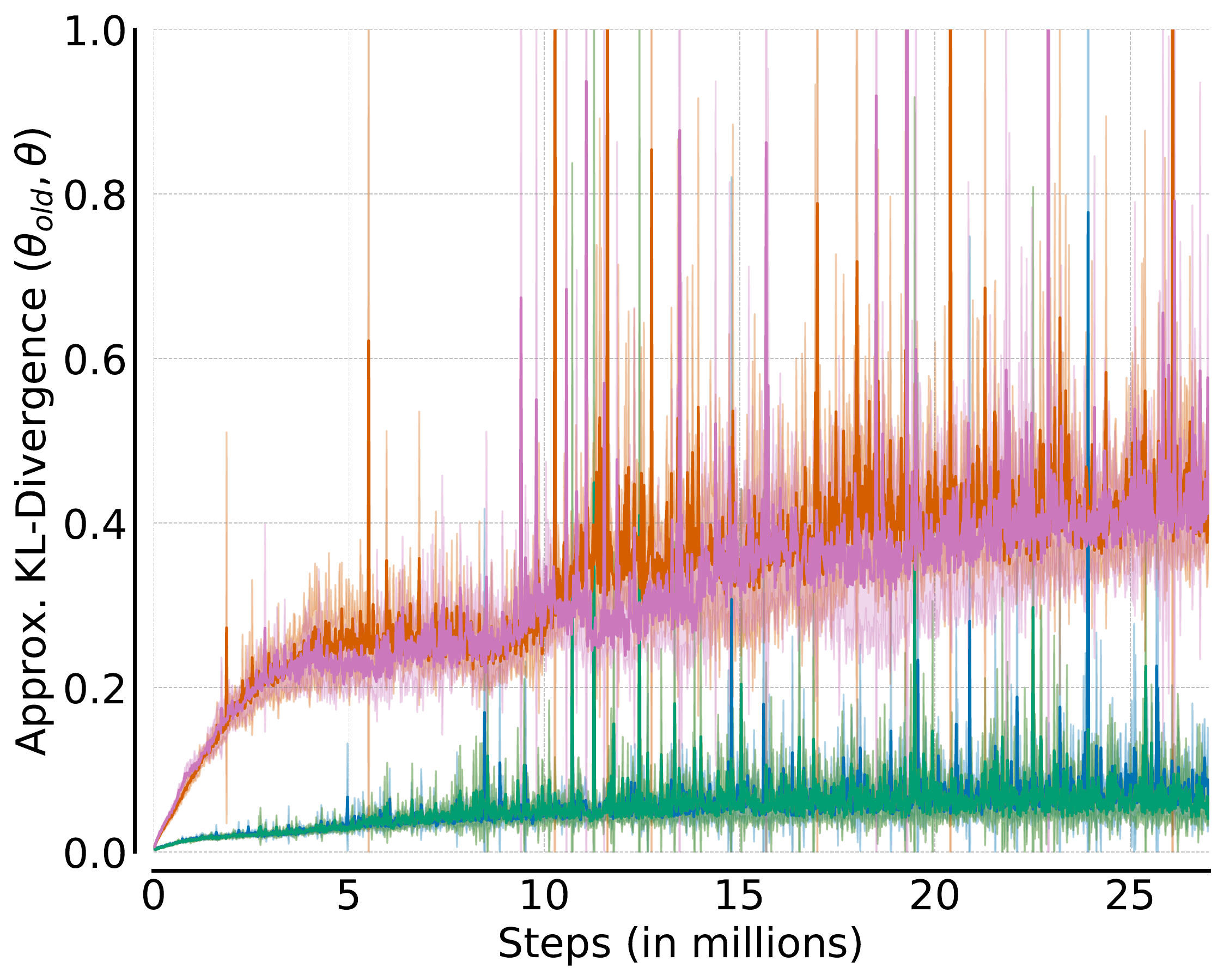}}
    \caption{Monitoring of the training runs in Dancing the Ballet when refreshing stale hidden states and advantages. a) shows the gradient norm of all model parameters, whereas b) presents the gradient norm only for parameters of the recurrent layer GRU. The Kullback–Leibler divergence between $\theta_{old}$ and $\theta$ is illustrated by c). All curves show the mean and the standard deviation.}
    \label{fig:ballet_refresh_analysis}
\end{figure}

We showed that refreshing stale hidden states and advantages do not yield any improvement concerning the IQM return (Section \ref{sec:refresh}.
When examining further training metrics (Figure \ref{fig:mm_refresh_analysis} and \ref{fig:ballet_refresh_analysis}), no substantial divergence between not refreshing and refreshing every or every other epoch is observed.
Gradient norms and the Kullback–Leibler divergence behave differently, while conducting 8 epochs of optimization.
Note that gradient norms are clipped at $0.5$.

\clearpage
\section{Adding the Last Action and Last Reward to the Observation}

\begin{figure}[hp!]
    \centering
    \subfigure[Mortar Mayhem Training Seeds]{\label{fig:mm_train}\includegraphics[width=\textwidth*\real{0.45}]{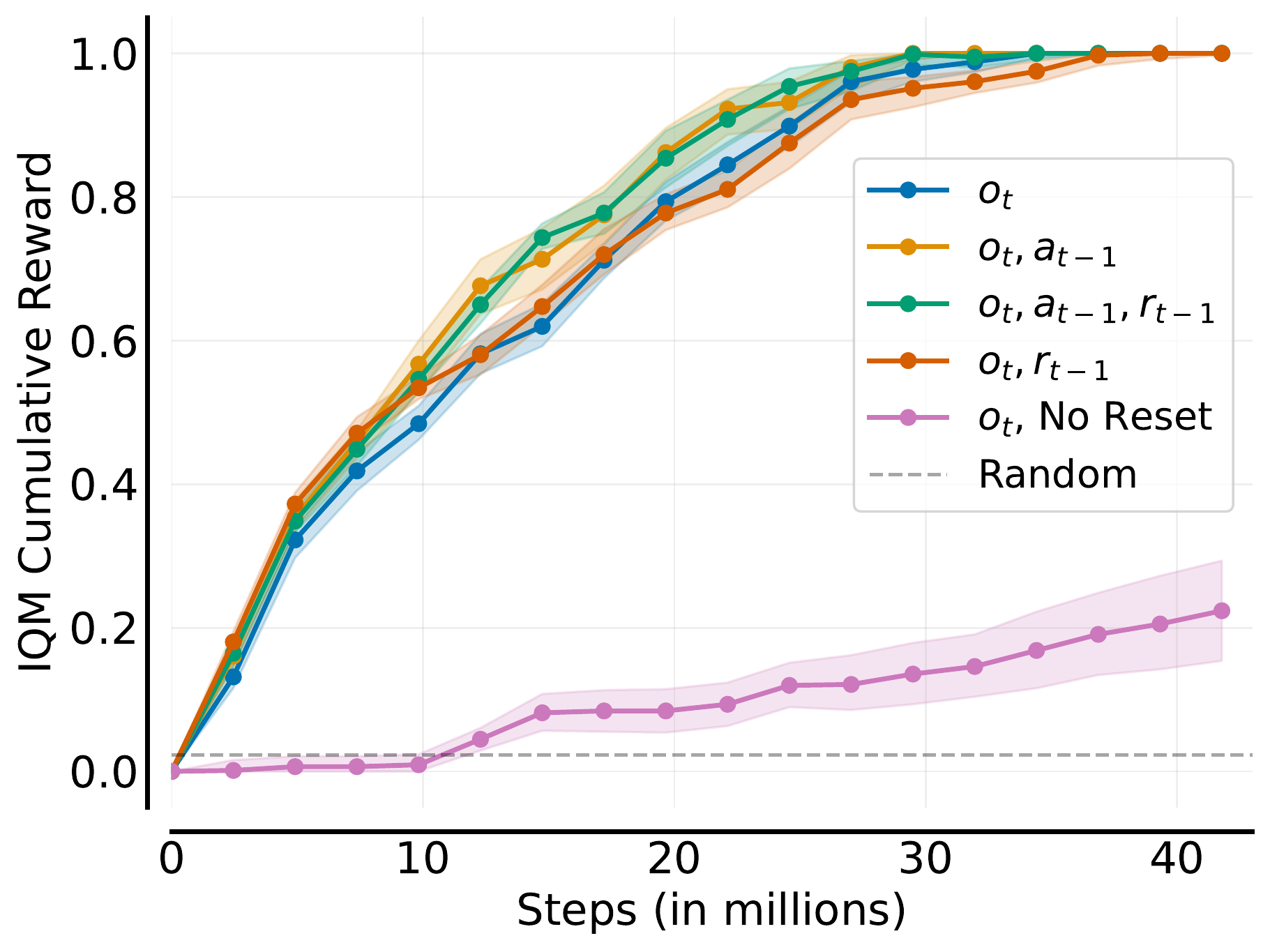}}
    \subfigure[Mortar Mayhem Novel Seeds]{\label{fig:mm_novel}\includegraphics[width=\textwidth*\real{0.45}]{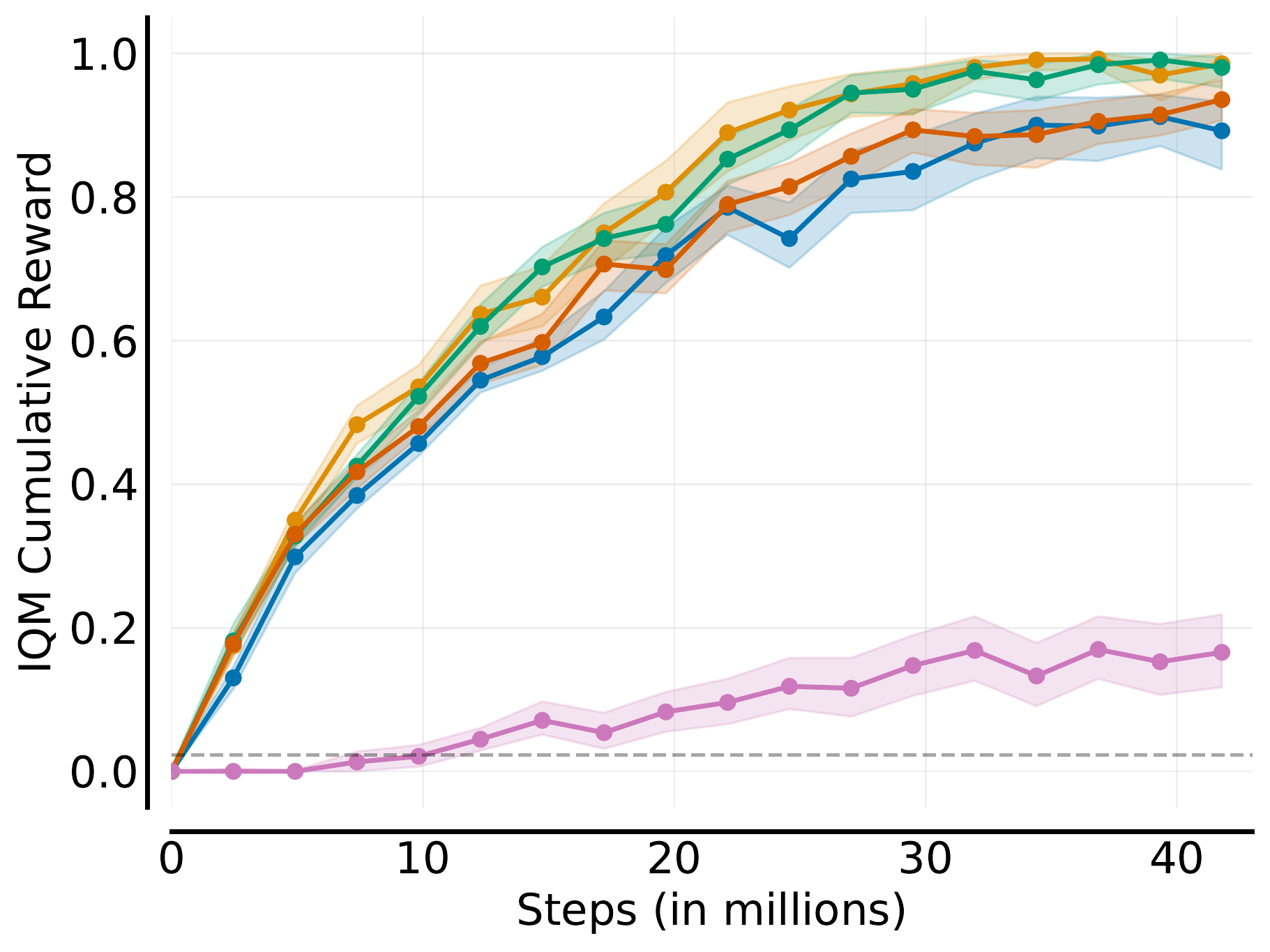}}
    \subfigure[Ballet 8 Dances Training Seeds]{\label{fig:ballet_train_8}\includegraphics[width=\textwidth*\real{0.9}]{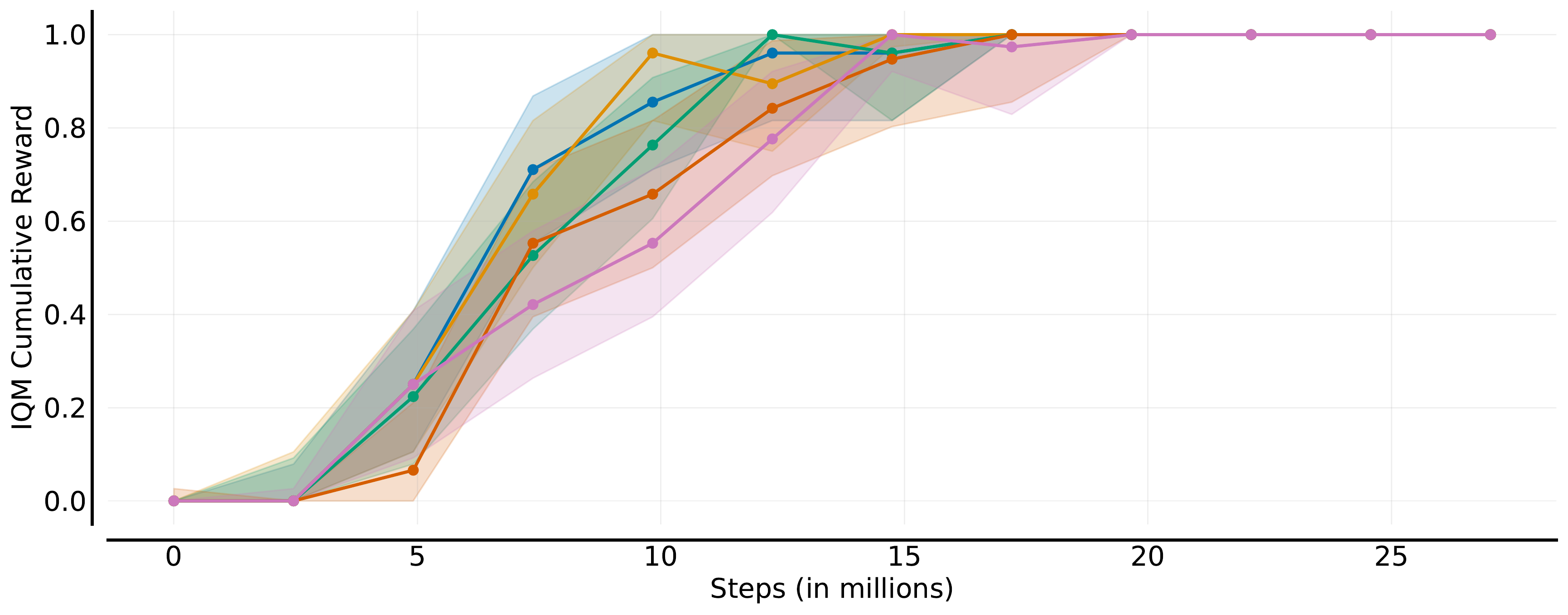}}
    \caption{An ablation on the agent's observation space. Also, the experiment on not resetting the hidden state for new episodes is shown.}
    \label{fig:experiments}
\end{figure}

Figure \ref{fig:experiments} describes an ablation on the agent's observation space to determine the significance of adding the last action, last reward or both to the current observation.
Of utility is usually the last action, whereas the last reward does not seem to affect performance.
Because of no notable extra cost, the last reward and last action can always be considered for the current observation regardless of their individual impact.
The figures also show an experiment when hidden states are not reset to zero.
Hence, the agent starts a new episode using the final hidden state of the former episode.

\end{document}